\documentclass[lettersize,journal]{IEEEtran}
\usepackage{amsmath,amsfonts}
\usepackage{algorithmic}
\usepackage{algorithm}
\usepackage{array}
\usepackage[caption=false,font=normalsize,labelfont=sf,textfont=sf]{subfig}
\usepackage{textcomp}
\usepackage{stfloats}
\usepackage{url}
\usepackage{verbatim}
\usepackage{graphicx}
\usepackage{cite}
\usepackage{hyperref}
\usepackage{url}
\usepackage{booktabs}       
\usepackage{amsfonts}       
\usepackage{nicefrac}       
\usepackage{microtype}      

\usepackage{graphicx}
\usepackage{arydshln}
\usepackage{booktabs}
\usepackage{multirow}
\usepackage{caption}
\usepackage{subcaption}
\usepackage{makecell}
\usepackage{csquotes}
\usepackage{epigraph}
\usepackage{amsmath}
\usepackage{cleveref}
\usepackage{pifont}
\usepackage{listings}
\usepackage{amssymb}
\usepackage{wrapfig}
\usepackage{tcolorbox}

\usepackage{colortbl}

\definecolor{lightpurple}{RGB}{230, 230, 250}

\newcommand\our{M4U}

\begin{document}

\title{\our{}: Evaluating Multilingual Understanding and Reasoning \\ for Large Multimodal Models}

\author{
    Hongyu Wang,
    Jiayu Xu,
    Senwei Xie,
    Ruiping Wang,~\IEEEmembership{Senior Member,~IEEE,} \\
    Jialin Li,
    Zhaojie Xie,
    Bin Zhang,
    Chuyan Xiong,
    Xilin Chen,~\IEEEmembership{Fellow,~IEEE}
    \thanks{The authors are with the Key Laboratory of AI Safety of Chinese Academy of Sciences (CAS), Institute of Computing Technology, CAS, Beijing 100190, China. Corresponding to: Ruiping Wang (wangruiping@ict.ac.cn), Xilin Chen (xlchen@ict.ac.cn)}
}

\markboth{Journal of \LaTeX\ Class Files,~Vol.~14, No.~8, August~2021}%
{Shell \MakeLowercase{\textit{et al.}}: A Sample Article Using IEEEtran.cls for IEEE Journals}


\maketitle

\begin{abstract}
Multilingual capability is an essential aspect for large multimodal models, since they are usually deployed across various countries and languages. However, most existing benchmarks for multilingual multimodal reasoning struggle to differentiate between models of varying performance; even language models without visual capabilities can easily achieve high scores. This leaves a comprehensive evaluation of leading multilingual multimodal models largely unexplored. In this work, we introduce \our{}, a novel and challenging benchmark for assessing the capability of multi-discipline multilingual multimodal understanding and reasoning. \our{} contains 10k samples covering 64 disciplines across 16 subfields in Science, Engineering, and Healthcare in six languages. Using \our{}, we conduct extensive evaluations of leading Large Multimodal Models (LMMs) and Large Language Models (LLMs) with external tools. The evaluation results demonstrate that the state-of-the-art model, GPT-4o, achieves only 47.6\% average accuracy on \our{}. Additionally, we observe that the leading LMMs exhibit significant language preferences. Our in-depth analysis indicates that leading LMMs, including GPT-4o, struggle to perform reasoning using multilingual information present in both visual and textual context. Specifically, they suffer performance degradation when prompted with cross-lingual multimodal questions. Our code and dataset is public available at \url{https://m4u-benchmark.github.io/m4u.github.io/}. 
\end{abstract}


\section{Introduction}

\IEEEPARstart{M}{ultimodal} reasoning is an essential aspect of human-level intelligence. AI systems with strong multimodal reasoning capabilities have extensive applications, including automatic scientific discovery, autonomous driving, and healthcare. The rapid advancements in Large Language Models (LLMs)~\cite{gpt4, llama, mistral} have led to the development of Large Multimodal Models (LMMs)~\cite{gemini, llavanext, deepseekvl, yi, qwen, qwen2vl} which demonstrate remarkable performance across a broad range of tasks, such as image captioning and visual question answering. Numerous benchmarks have been established to comprehensively evaluate these leading LMMs in real-world scenarios~\cite{MMBench, seedbench, pope, visualwebarena, mme}. Unlike perceptual tasks~\cite{vqav2, coco}, multimodal reasoning tasks, including mathematical reasoning~\cite{mathvista} and scientific question answering~\cite{scienceqa, ai2d}, present significant challenges for neural models. These tasks necessitate an understanding of domain-specific knowledge and the ability to perform complex logical reasoning alongside visual content. Additionally, multilingual capability is crucial for real-world applications, as these models are typically deployed across various countries and languages.
 
Many datasets are curated to evaluate the capability of multilingual multimodal reasoning. However, the multimodal component of existing benchmarks~\cite{m3exam, examsv, cmb} is limited in scale. We observe that the current data on multilingual multimodal reasoning suffers from language disparities in task complexity. For instance, the multimodal part of M3Exam~\cite{m3exam} contains 61\% high-difficulty questions in English, but only 23\% high-difficulty questions in Chinese. Furthermore, the existing benchmark struggles to differentiate between models of varying multimodal capabilities. As shown in Table~\ref{tab:benchmark}, without any visual information, the state-of-the-art multilingual LLM easily achieves high scores on the Chinese and English sections of M3Exam. Consequently, a systematic evaluation of multilingual multimodal understanding and reasoning for leading models remains largely unexplored.

\begin{figure}
    \centering
    \includegraphics[width=\linewidth]{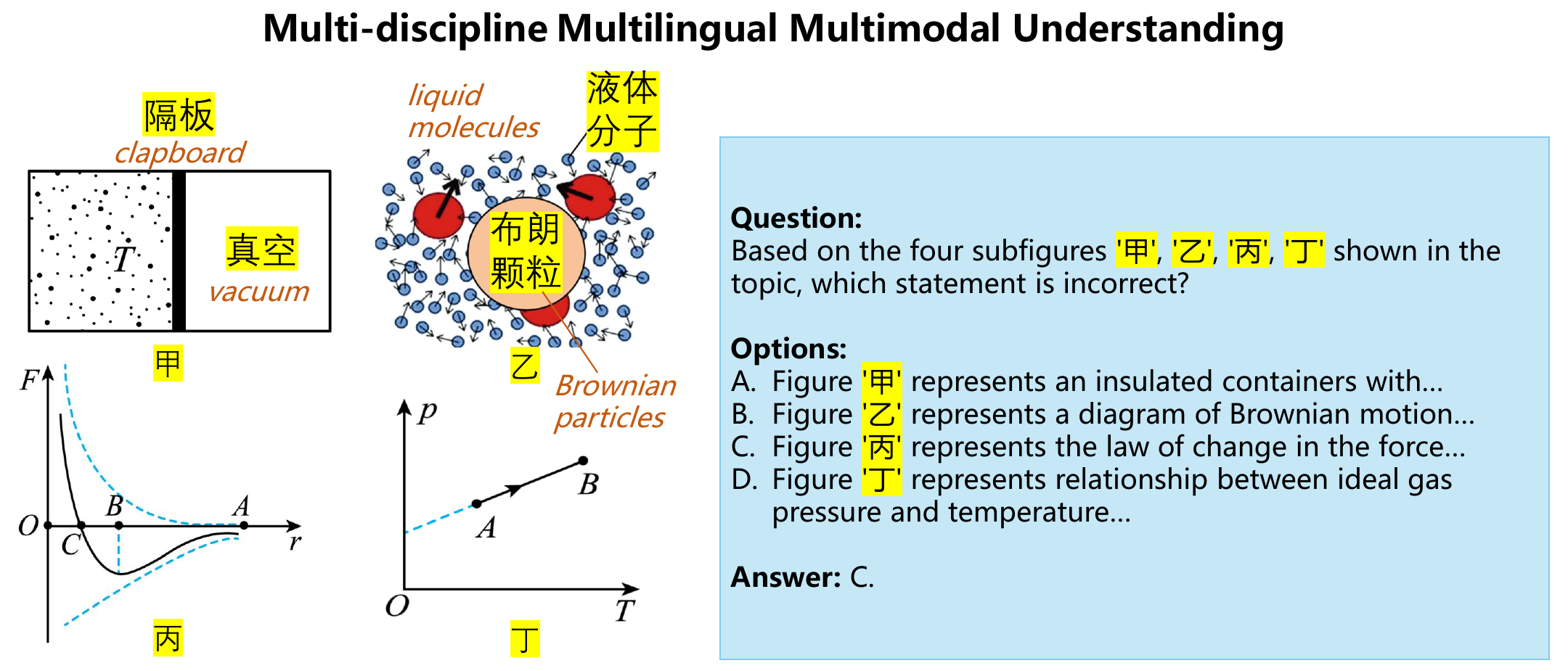}
    \caption{An illustration of multi-discipline multilingual multimodal understanding. Both textual questions and images contain the multilingual contents. We highlight the Chinese contents in yellow. English translations are provided for better readability.}
    \label{fig:abstract}
\end{figure}

\begin{figure*}[t]
    \centering
    \begin{minipage}{0.5\textwidth}
        \includegraphics[width=\linewidth]{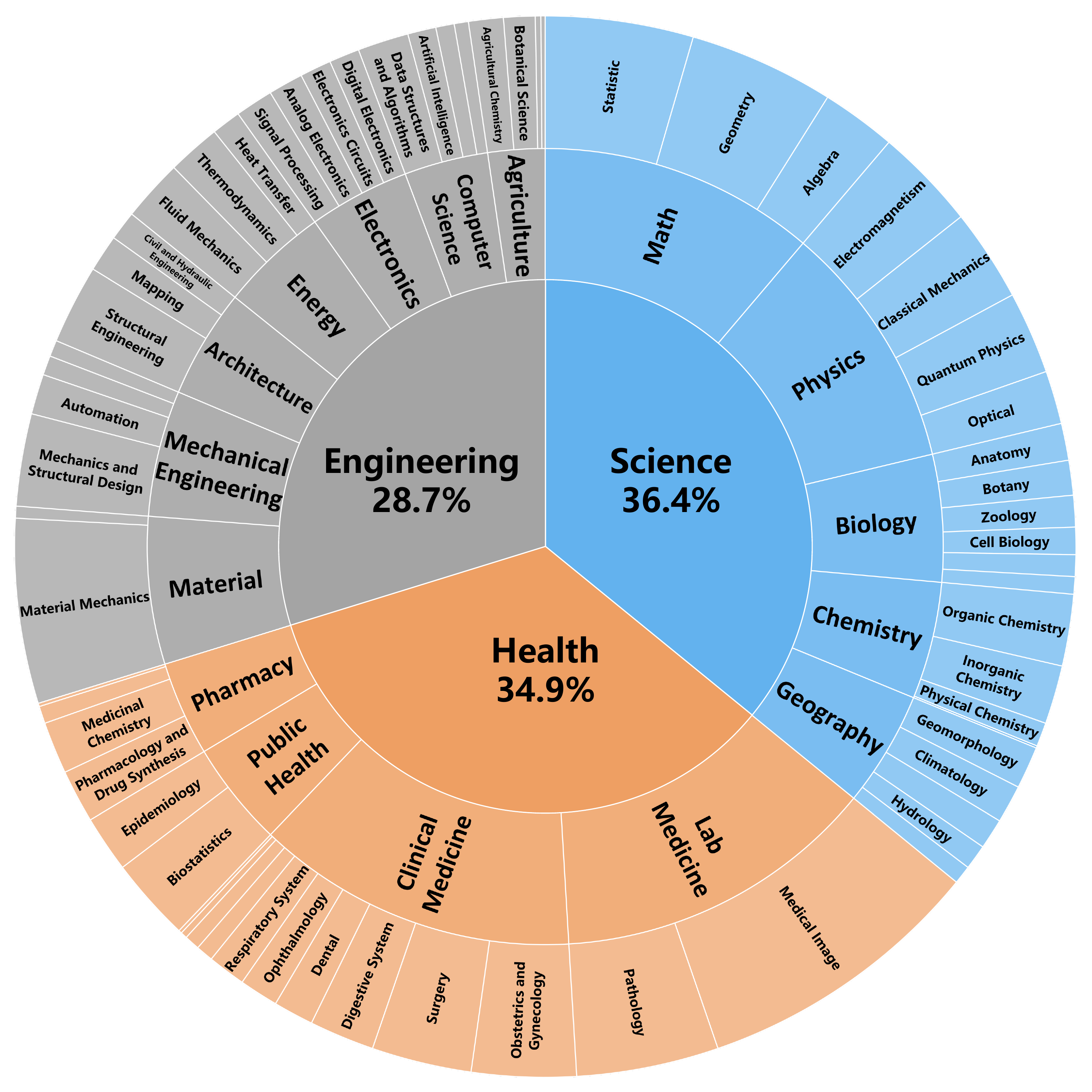}
    \end{minipage}\hfill
    \begin{minipage}{0.5\textwidth}
        \centering
        \begin{tabular}{l|cc}
            \toprule
            \bf Statistics  & \bf test & \bf test-mini \\
            \midrule
            Languages & ZH, EN, DE & \makecell{ZH, EN, DE, \\ JA, AR, TH} \\
            \midrule
            Total questions & 8,931  & 1,074 \\
            Total disciplines / subfields & \multicolumn{2}{c}{64 / 16} \\
            Total image types & \multicolumn{2}{c}{13} \\
            \midrule
            Image in the question & 8,271 & 1008 \\
            $\quad$Image at the beginning & 6,205 & 744\\
            $\quad$Image in the middle & 1,321 & 204 \\
            $\quad$Image at the end & 745 & 60\\
            Image in the options & 660 & 66 \\
            \midrule 
            Single / multiple image(s) & 8,199 / 732 & 996 / 78\\
            \midrule
            Maximum question length & 279 & 114 \\
            Maximum option length &  63 & 33 \\
            Average question length & 33.2 & 31.5 \\
            Average option length & 6.1 & 6.0 \\
            \bottomrule
        \end{tabular}
    \end{minipage}
    \caption{Key statistics of \our{} dataset. \our{} covers a wide scope of tasks from Science, Engineering and Health in Chinese, English and German, and supports the interleaved vision-language documents.}
    \label{fig:stats}
\end{figure*}

\begin{figure*}[t]
      \centering
      \includegraphics[width=0.95\textwidth]{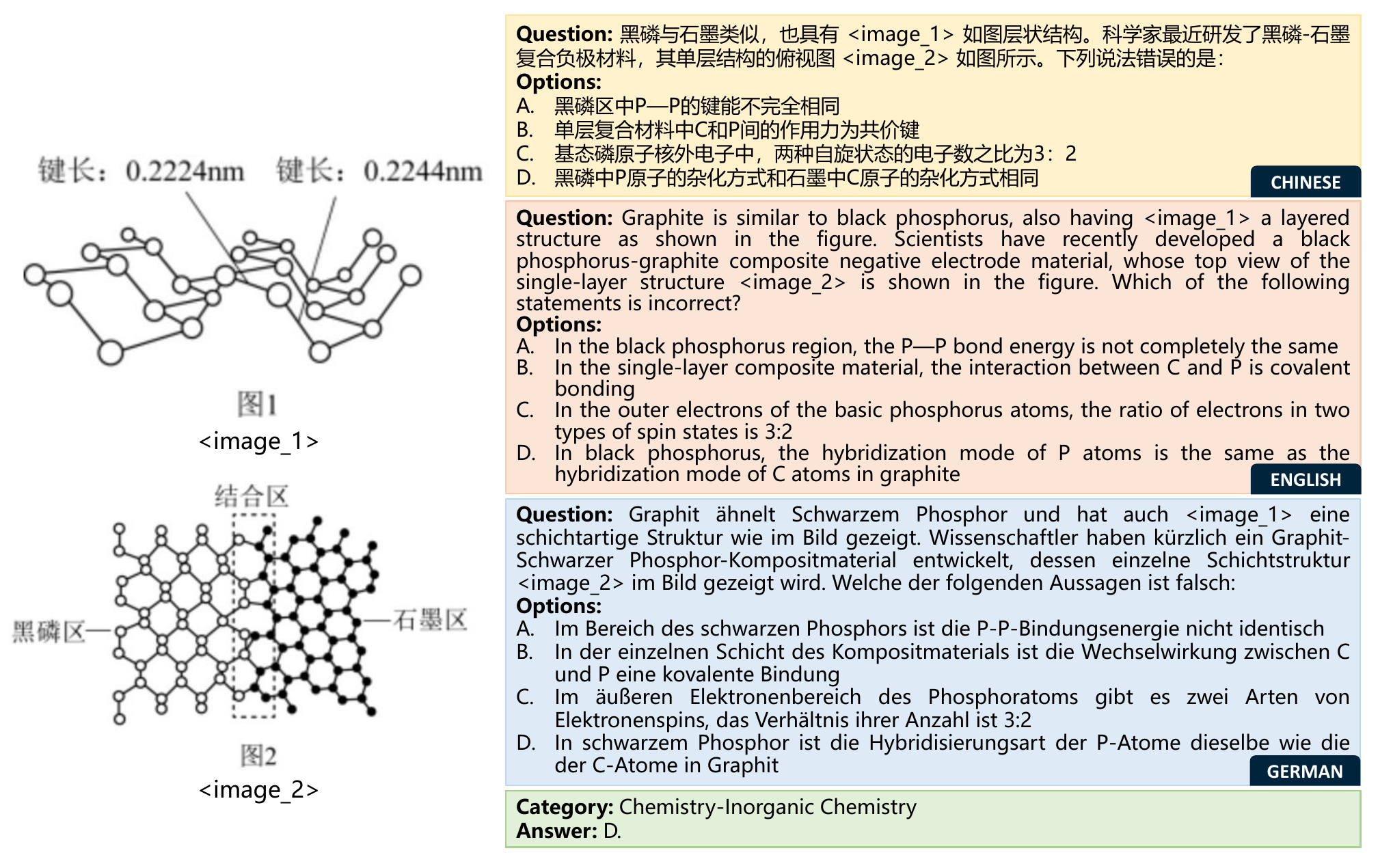}
      \caption{An example from the Chemistry-Inorganic of \our{} dataset. The sample contains multiple images, and has multilingual contents in the question and images.}
      \label{fig:samples}
\end{figure*}

To advance the development of multilingual LMMs, we introduce \our{}, a novel and challenging benchmark for evaluating foundational models on expert-level multilingual multimodal understanding and reasoning. Specifically, we assembled a team of more than 10 college and graduate students to collect a high-quality dataset and assess its difficulty and correctness. As shown in Figure~\ref{fig:stats}, \our{} consists of 10,005 multiple-choice questions covering 64 disciplines across 16 subfields in Science, Engineering, and Healthcare. To minimize the risk of data contamination, samples are collected from college exams and quizzes from online video lectures. Additionally, a significant portion (35\%) of the questions in \our{} are written by our team based on textbooks. Figure~\ref{fig:samples} illustrates an example from the Chemistry-Inorganic part of \our{}, demonstrating that our dataset requires expert-level multimodal reasoning and multilingual capability.

With \our{}, we conduct a comprehensive evaluation, both quantitative and qualitative, on the zero-shot performance of 22 leading LMMs and 4 LLMs. Furthermore, we assess the performance of the LMMs with chain-of-thought prompting~\cite{cot, mmcot} and the LLMs with external tools, such as a powerful captioning model. As shown in Table~\ref{tab:zs} in \textsection \ref{sec:main}, the most advanced model, GPT-4o~\cite{gpt4o}, achieves only 47.6\% average accuracy with zero-shot prompting on the \our{} dataset, demonstrating the significant challenge \our{} poses for existing models. Additionally, we observe significant language preferences among the leading LMMs: InstructBLIP Vicuna-7B achieves 29.8\% accuracy on the English section, but only 13.7\% and 19.7\% accuracy on the Chinese and German sections, respectively. Further results (\textsection \ref{sec:cross_lingual}) shows that leading LMMs suffer performance degradation when prompted with cross-lingual multimodal questions, such as images with key textual information in Chinese while the question is in German. These results indicate that existing models still struggle to perform complex reasoning based on multilingual content in visual and textual data. Our in-depth analysis (\textsection \ref{sec:case_study}) reveals that the errors of GPT-4V(ision) are mainly due to limited perception ability, domain-specific knowledge, and reasoning. These findings demonstrate that LMMs still have significant room for improvement, particularly in multilingual multimodal reasoning.

\section{Related Work}
Recent years have witnessed a trend towards large-scale multimodal pre-training, which aims to unify various vision-language tasks with a single model~\cite{instructblip, deepseekvl, qwenvl, gpt4, gpt4o}. With the rapid progress of LMMs, previous benchmarks (e.g., VQA-v2~\cite{vqav2}, GQA~\cite{gqa}) are insufficient to comprehensively evaluate the general multimodal capability of these models. Therefore, many datasets are curated for evaluate different aspects of multimodal capability, spanning from robustness and hallucination (e.g., POPE~\cite{pope}) to general perception capability (e.g., MM-Vet~\cite{mmvet} and MMBench~\cite{MMBench}). As for multimodal reasoning tasks, MathVista~\cite{mathvista} presented a collection of diverse challenging mathematical and visual tasks. After that, instead focusing on the mathematical domain, MMMU~\cite{mmmu} introduced a large-scale collection of more difficult expert-level problems that cover 30 different subjects. However, these benchmarks are primarily focused on English. 

Multilingual capability is crucial for LLMs and LMMs. Many holistic evaluations have been conducted for LLMs, such as PAWS-X~\cite{pawsx}, XCOPA~\cite{xcopa}, XStoryCloze~\cite{xstorycloze}, MGSM~\cite{mgsm}, BIG-bench~\cite{bigbench}, MMMLU~\cite{mmmlu}, Global MMLU~\cite{global_mmlu}. As for the evaluation of LMMs, M3Exam~\cite{m3exam} collects the official exam papers of 9 different languages. However, they mainly focus on the evaluation of language capability. Despite M3Exam contains the samples with image as the input, its multimodal part is limited in scale. Besides, M3Exam struggles to differentiate between models of varying multimodal performance. As shown in Table~\ref{tab:benchmark}, a language model without visual capabilities can easily achieve high scores. Different from previous works, \our{} covers over 64 disciplines across Science, Engineering and Healthcare. We conduct the strict collection guidelines and quality control, which ensures that \our{} requires significant visual efforts and domain-specific knowledge to perform multi-step reasoning.

\section{The \our{} Benchmark}
\label{sec:m4u}

\subsection{Overview}

In this section, we introduce \our{}, a novel and challenging benchmark to assess multilingual multimodal understanding and reasoning of foundational models. To investigate whether there are differences in the multimodal reasoning capabilities of LMMs in different languages, we first construct the Chinese part of \our{} and then translate it into English and German. This approach ensures that the domain-specific knowledge and reasoning abilities tested in different languages remain consistent. Specifically, we assembled a team of more than 10 college students to collect questions from the Internet, textbooks, online video lectures, and college exams. Subsequently, a team of graduate students from related majors assessed the quality of the curated questions. Following this, we utilize GPT-4 Turbo (\texttt{gpt-4-turbo-preview}) to translate the questions into other languages and then manually check the quality of the translated questions. 

To support more medium- or low-resource languages, we present \our{}-mini, a tiny version of M4U with three additional languages (Japanese, Arabic, and Thai). We randomly select 5\% of the test data and follow the same processing pipeline to construct these parts.

The key statistics of \our{} are detailed in Figure~\ref{fig:stats}. \our{} contains 10,005 multiple-choices questions, covering 64 subjects of 16 fields from Science (36.4\%), Engineering (28.7\%) and Health (34.9\%). Different from the prior work~\cite{m3exam}, \our{} includes interleaved image-text documents: 8.2\% of the questions have multiple images, while the images of 14.8\% and 7.4\% of the questions are placed in the middle of question and the options, respectively. The average length of questions and options is 33.2 words and 6.1 words. The image sources of \our{} cover 13 categories in different scenarios, e.g., diagrams, technical blueprints and medical images. 

\begin{table*}[t]
    \centering
    \small
    \setlength{\tabcolsep}{12pt}
    \caption{Comparison between \our{} and the existing benchmarks. $^*$We report the size of test set of the multimodal part for the benchmarks. }
    \begin{tabular}{l|ccccl}
    \toprule
    \bf Benchmark & \bf Multilingual & \bf Multimodal & \bf Size$^*$ & \bf Difficulty & \bf Fields \\
    \midrule
    CMMLU~\cite{cmmlu} & \ding{55} & \ding{55} & - & \ding{72} \ding{72} \ding{72} & STEM, Humanities, etc. \\
    C-Eval~\cite{ceval} & \ding{55} & \ding{55}  & - & \ding{72} \ding{72} \ding{72} & STEM, Humanities, etc.\\
    MMLU~\cite{mmlu} & \ding{55} & \ding{55} & - & \ding{72} \ding{72} \ding{72} & STEM, Humanities, etc. \\
    \midrule
    MathVista~\cite{mathvista} & \ding{55} & \ding{51} &  6,141 & \ding{72} \ding{72} \ding{72} & Mathematics \\
    CMMMU~\cite{cmmmu} & \ding{55} & \ding{51} & 12,012 & \ding{72} \ding{72} \ding{72} & Art, STEM, Humanities, etc. \\
    MMMU~\cite{mmmu} & \ding{55} & \ding{51} & 11,550 & \ding{72} \ding{72} \ding{72} & Art, STEM, Humanities, etc. \\
    \midrule
    MGSM~\cite{mgsm} & \ding{51} & \ding{55} & - &  \ding{72} \ding{72} \ding{72}  & Mathematics \\
    EXAMS-V~\cite{examsv} & \ding{51} & \ding{51} & 1,221 & \ding{72} \ding{72} & STEM, Humanities, etc. \\
    M3Exam~\cite{m3exam} & \ding{51} & \ding{51} & 2,816 & \ding{72} & STEM, Humanities, etc. \\
    \midrule
    \our{} (ours) & \ding{51} & \ding{51} & 10,005 & \ding{72} \ding{72} \ding{72} \ding{72} & STEM, Healthcare \\
    \bottomrule
    \end{tabular}
    \label{tab:comparison}
\end{table*}

\begin{table*}[t]
    \centering
    \small
    \setlength{\tabcolsep}{12pt}
    \caption{The zero-shot accuracy of the state-of-the-art multilingual LLMs on the multimodal part of M3Exam and \our{} dataset. We provide the scores of random choices in \textcolor{blue}{blue} as the reference baseline. The higher scores of the text-only LLM indicate that the multimodal benchmark requires less visual efforts.}
    \begin{tabular}{l|cccccc}
    \toprule
    \bf Benchmark & \bf Chinese$\downarrow$ & \bf English$\downarrow$ & \bf German$\downarrow$ & \bf Thai$\downarrow$ & \bf Japanese$\downarrow$ & \bf Arabic$\downarrow$ \\
    \midrule
    \multicolumn{6}{l}{\ \emph{Qwen1.5-14B-Chat}} \\
    M3Exam~\cite{m3exam} & 66.4 \textcolor{blue}{(25.9)} & 56.0 \textcolor{blue}{(25.0)} & - & 25.2 \textcolor{blue}{(22.9)} & - & - \\
    \our{} (ours) & 28.0 \textcolor{blue}{(25.9)} & 19.7 \textcolor{blue}{(25.9)} & 27.6 \textcolor{blue}{(25.9)} & - & - & - \\
    \our{}-mini (ours) & 28.9 \textcolor{blue}{(26.3)} & 17.7 \textcolor{blue}{(26.3)} & 29.5 \textcolor{blue}{(26.3)} & 19.3 \textcolor{blue}{(26.3)} & 26.9 \textcolor{blue}{(26.3)} & 12.0 \textcolor{blue}{(26.3)} \\
    \midrule
    \multicolumn{6}{l}{\ \emph{Qwen2.5-14B-Instruct}} \\
    M3Exam~\cite{m3exam} & 67.8 \textcolor{blue}{(25.9)} & 63.0 \textcolor{blue}{(25.0)} & - & 34.7 \textcolor{blue}{(22.9)} & - & - \\
    \our{} (ours) &  30.4 \textcolor{blue}{(25.9)} & 23.7 \textcolor{blue}{(25.9)} & 22.5 \textcolor{blue}{(25.9)} & - & - & - \\
    \our{}-mini (ours) & 35.0 \textcolor{blue}{(26.3)} & 25.7 \textcolor{blue}{(26.3)} & 25.7 \textcolor{blue}{(26.3)} & 13.6 \textcolor{blue}{(26.3)} & 35.7 \textcolor{blue}{(26.3)} & 13.8 \textcolor{blue}{(26.3)} \\
    \bottomrule
    \end{tabular}
    \label{tab:benchmark}
\end{table*}

We present a comparison of \our{} with existing benchmarks in Table~\ref{tab:comparison}. Unlike MMMU~\cite{mmmu} and CMMMU~\cite{cmmmu}, our dataset focuses on the evaluation of multilingual multimodal reasoning. Furthermore, \our{} is larger and has a more balanced distribution of difficulty across different languages compared to M3Exam~\cite{m3exam}. This ensures a fair comparison of models' capabilities in multimodal reasoning within multilingual scenarios. More importantly, we implement strict collection guidelines and quality control measures to minimize the risk of data contamination. To quantitatively measure the difficulty on visual capability for these benchmarks, we use the scores of LLMs without any visual information as a reference. Higher scores indicate that the benchmark is less effective at differentiating between models of varying multimodal performance. As shown in Table~\ref{tab:benchmark}, for the evaluation on Chinese and English, without any visual information, Qwen2.5-14B-Instruct easily achieves over 60\% accuracy on M3Exam, while it only has 27.0\% and 30.4\% average accuracy on test and test-mini set of our dataset, respectively. This suggests that \our{} is more challenging and less exposed to the training corpus of LLMs. Detailed dataset document can be found in supplementary materiel.

\subsection{Data Collection}

\textbf{Data sources.} Following MMMU~\cite{mmmu}, we go through the educational programs of top universities, then select 64 subjects of 16 subfields from Science, Engineering and Healthcare whose applications highly rely on visual information. We recruit a team of over 10 college students to collect multiple-choices questions from public available sources. To minimize the risk of data contamination for foundation models, different from M3Exam~\cite{m3exam}, we do not include the samples from the official exam papers, e.g., National Postgraduate Entrance Examination and national professional exams. Although these resources usually have higher quality and are well organized, they are also easy to be curated for the training of LLMs. Therefore, we carefully select the data sources for \our{}: most questions of our dataset are collected from the quizzes of online video lectures and college exams in PDF documents. The guidelines for annotators stress the importance of strictly following copyright and licensing rules from the original data sources, particularly avoiding materials from websites that prohibit copying and redistribution.

\textbf{Data processing.} The primary sources of \our{} include college exams, the quizzes of online video lectures and the written questions. Most of college exams are uploaded by their students as images or scanned PDF documents, while the quizzes of online video lectures can be taken as the screenshot. We first adopt the OCR tools to convert these images into plain texts, then manually correct the potential errors of OCR results. Besides, we also write a large portion (35\%) of questions according to the textbooks. For the mathematical formulas and the chemical structures, we require the annotators to convert them into \LaTeX{} format. Since the samples of \our{} may include multiple images in the questions or options, the annotators also annotate the location and type of each image (e.g. tables, blueprints and medical images). 

After collecting the data, we design a two-stage post-processing pipeline to further improve the quality of \our{}. We first design the guidelines to allow each annotator to score the collected samples from three dimensions: image quality, question description quality and the difficulty of visual understanding, and filter out the questions with average scores lower than 2.0. Then we recruit a team of graduate students of related major to assess the difficulty and quality of the curated questions. We further filter out the questions with the minimum visual efforts and the wrong answer. After that, we use GPT-4 Turbo to translate the Chinese part of \our{} to other languages. Then the annotators will check and correct the potential errors introduced by machine translation.

To ensure the integrity and reliability of the data, we have established a rigorous quality control process for multilingual content. Annotators are instructed to leverage translation tools to back-translate text from unfamiliar languages into either Chinese or English. This step enables a thorough verification of the semantic accuracy, contextual alignment, and consistency of the translations. For domain-specific terminology, annotators are required to perform meticulous manual reviews, cross-referencing authoritative resources to ensure correctness and adherence to specialized contexts.

\subsection{Evaluation}

\textbf{Instruction design.} We evaluate the zero-shot performance of 22 leading LMMs of different scales on \our{}. To minimize the format discrepancy between training and evaluation, we handle models that support interleaved image-text documents by inserting the visual tokens of each image into the corresponding position as in training. For models that only support image-text pairs as input, we place all visual tokens at the beginning of the sentence and use annotated positions to refer to each image. Furthermore, we also evaluate the performance of various LMMs with chain-of-thought prompting~\cite{cot, mmcot} and LLMs equipped with detailed visual captions.

\begin{tcolorbox}[colback=blue!5!white, colframe=blue!75!black, title=Evaluation template to extract rationales]
\textbf{\{Question\}} \\
A. \textbf{\{1st Option\}} \\
B. \textbf{\{2nd Option\}} \\
C. \textbf{\{3rd Option\}} \\
D. \textbf{\{4th Option\}} \\
Please analyze the question and options.
\end{tcolorbox}

\begin{tcolorbox}[colback=blue!5!white, colframe=blue!75!black, title=Evaluation template for English]
\textbf{\{Question\}} \\
A. \textbf{\{1st Option\}} \\
B. \textbf{\{2nd Option\}} \\
C. \textbf{\{3rd Option\}} \\
D. \textbf{\{4th Option\}} \\
\textbf{\{Generated Chain-of-Thought\} (Optional)} \\
Answer with the option's letter from the given choices directly.
\end{tcolorbox}

We present the prompt template used for the zero-shot evaluation of \our{}. For chain-of-thought prompting, we first prompt the model to generate a rationale for the question and the options, which is then appended after the options. Finally, the models are required to follow the instruction to directly generate the predicted option for each question.

\textbf{Overall results.} As shown in Table~\ref{tab:zs}, the existing models still lack the capability for expert-level multilingual multimodal understanding and reasoning: the most advanced GPT-4o achieves only 47.6\% average accuracy with zero-shot prompting on the \our{} dataset. Additionally, we observe that these models exhibit strong language preferences. For instance, InstructBLIP Vicuna-7B achieves 29.8\% accuracy on the English part, but only 13.7\% and 19.7\% accuracy on the Chinese and German parts, respectively. These results indicate that there remains significant room for improvement in LMMs, particularly regarding multilingual capability and complex multimodal reasoning.

\begin{table*}[t]
    \setlength{\tabcolsep}{12pt}
    \centering
    \small
    \caption{The zero-shot accuracy of various LLMs, augmented LLMs and LMMs on \our{} dataset. \textit{CoT} is short for chain-of-thought prompting.}
    \begin{tabular}{l|l|cccc}
    \toprule
    \bf Models & \bf Size & \bf Chinese$\uparrow$ & \bf English$\uparrow$ & \bf German$\uparrow$ & \bf Average$\uparrow$ \\
    \midrule
    Random choices & - & \multicolumn{4}{c}{25.9}  \\
    \midrule
    \multicolumn{6}{l}{\ \emph{Large Language Models}} \\
    Qwen1.5-7B-Chat~\cite{qwen} & 7B & 29.5 & 15.0 & 28.5 & 24.3\\
    Qwen1.5-14B-Chat~\cite{qwen} & 14B & 28.0 & 19.7 & 27.6 & 25.1\\
    \midrule
    \multicolumn{6}{l}{\ \emph{Augmented Large Language Models (+ Visual Caption)}} \\
    Mistral-Instruct-v0.2-7B~\cite{mistral} & 7B & 24.9 & 24.9 & 26.9 & 25.6\\
    Gemini 1.0 Pro~\cite{gemini} & - & 31.6 & 31.1 & 30.9 & 31.2 \\
    Qwen1.5-7B-Chat~\cite{qwen} & 7B & 34.2 & 27.7 & 31.7 & 31.2\\
    Qwen1.5-14B-Chat~\cite{qwen} & 14B & 32.7 & 32.0 & 33.8 & 32.8\\
    \midrule
    \multicolumn{6}{l}{\ \emph{Large Multimodal Models}} \\
    VisualGLM~\cite{visualglm} & 6B & 8.7 & 22.4 & 13.5 & 14.9\\
    Ying-VLM~\cite{yingvlm} & 13B & 22.3 & 11.2 & 15.6 & 16.4\\
    InstructBLIP-Vicuna-13B~\cite{instructblip} & 13B & 10.5 & 23.4 & 18.6 & 17.5\\
    InstructBLIP-Vicuna-7B~\cite{instructblip} & 7B & 13.7 & 28.1 & 19.7 & 20.5\\
    LLaVA-NeXT-Vicuna-7B~\cite{llavanext} & 7B & 11.8 & 29.8 & 28.2 & 23.3\\
    LLaVA-NeXT-Vicuna-13B~\cite{llavanext} & 13B & 21.9 & 30.9 & 29.3 & 27.4\\
    Qwen-VL-Chat~\cite{qwenvl} & 7B & 29.7 & 29.9 & 27.1 & 28.9\\
    CogVLM-Chat~\cite{cogvlm} & 7B & 28.9 & 30.2 & 28.5 & 29.2\\
    LLaVA-NeXT-Mistral-7B~\cite{llavanext} & 7B & 28.2 & 30.6 & 29.4 & 29.4\\
    InternLM-XComposer~\cite{internlmx} & 7B & 31.8 & 31.6 & 29.1 & 30.8\\
    DeepSeek-VL~\cite{deepseekvl} & 7B & 30.4 & 32.8 & 30.8 & 31.3\\
    Yi-VL-6B~\cite{yi} & 6B & 33.4 & 31.4 & 29.7 & 31.5\\
    Yi-VL-34B~\cite{yi} & 34B & 33.5 & 33.3 & 30.5 & 32.4\\
    Gemini 1.0 Pro~\cite{gemini} & - & 34.9 & 32.7 & 30.8 & 32.8 \\
    LLaVA-NeXT-34B~\cite{llavanext} & 34B & 38.5 & 36.2 & 35.2 & 36.6\\
    GPT-4V(ision)~\cite{gpt4} & - & 39.7 & 39.4 & 37.3 & 38.8 \\
    GPT-4o~\cite{gpt4o} & - &  \bf 49.4 & \bf 47.8 & \bf 45.6 & \bf 47.6 \\
    \midrule
    \multicolumn{6}{l}{\ \emph{ Augmented Large Multimodal Models}} \\
    Gemini 1.0 Pro~\cite{gemini} + \textit{CoT} & - & 34.4 & 34.2 & 33.9 & 34.2 \\
    GPT-4V(ision)~\cite{gpt4} + \textit{CoT} & - & \bf 43.9 & \bf 43.6 & \bf 40.3 & \bf 42.6 \\
    \bottomrule
    \end{tabular}
    \label{tab:zs}
\end{table*}

\begin{table*}[t]
    \setlength{\tabcolsep}{12pt}
    \centering
    \small
    \caption{The zero-shot accuracy of various LMMs, augmented LMMs on the cross-lingual set of \our{} dataset. \textit{CoT} is short for chain-of-thought prompting.}
    \begin{tabular}{l|c|cccc}
    \toprule
    \bf Models & \bf Size & \bf Chinese$\uparrow$ & \bf English$\uparrow$ & \bf German$\uparrow$ & \bf Average$\uparrow$ \\
    \midrule
    DeepSeek-VL~\cite{deepseekvl} & 7B & 32.8 & 34.0 & 33.3 & 33.4\\
    Yi-VL-6B~\cite{yi} & 6B & 39.2 & 34.3 & 30.1 & 34.5\\
    Gemini 1.0 Pro~\cite{gemini} & - & 38.0 & 36.3 & 32.9 & 35.7\\
    Yi-VL-34B~\cite{yi} & 34B & 41.6 & 38.7 & 34.2 & 38.2\\
    LLaVA-NeXT-34B~\cite{llavanext} & 34B & 44.6 & 40.9 & 36.1 & 40.5\\
    GPT-4V(ision)~\cite{gpt4} & - & 45.3 & 41.2 & 38.2 & 41.6 \\
    GPT-4o~\cite{gpt4o} & - & \bf 52.0 & \bf 47.5 & \bf 45.2 & \bf 48.2 \\
    \midrule
    Gemini 1.0 Pro~\cite{gemini} + \textit{CoT} & - & 38.1 & 35.7 & 37.8 & 37.2\\
    GPT-4V(ision)~\cite{gpt4} + \textit{CoT} & - & \bf 46.7 & \bf 48.0 & \bf 42.6 & \bf 45.8 \\
    \bottomrule
    \end{tabular}
    \label{tab:cross_lingual}
\end{table*}

\section{Experiments}
\label{sec:exp}

\subsection{Setup}

We evaluate the performance of zero-shot learning for various LMMs and LLMs of different scales across different languages on \our{} dataset. The models are prompted to directly generate the option's letter. Further we also evaluate the performance of the LMMs with chain-of-thought prompting~\cite{cot, mmcot}: the models should first generate the rationale for the question and the options, then give the predicted option. For reference, we add the baseline of Random choices: we randomly select an option, and use the average accuracy of 30 runs with different seeds. We adopt NVIDIA A40 with 48GB memory for evaluations.

\begin{figure*}[t]
    \centering
      \includegraphics[width=0.48\linewidth]{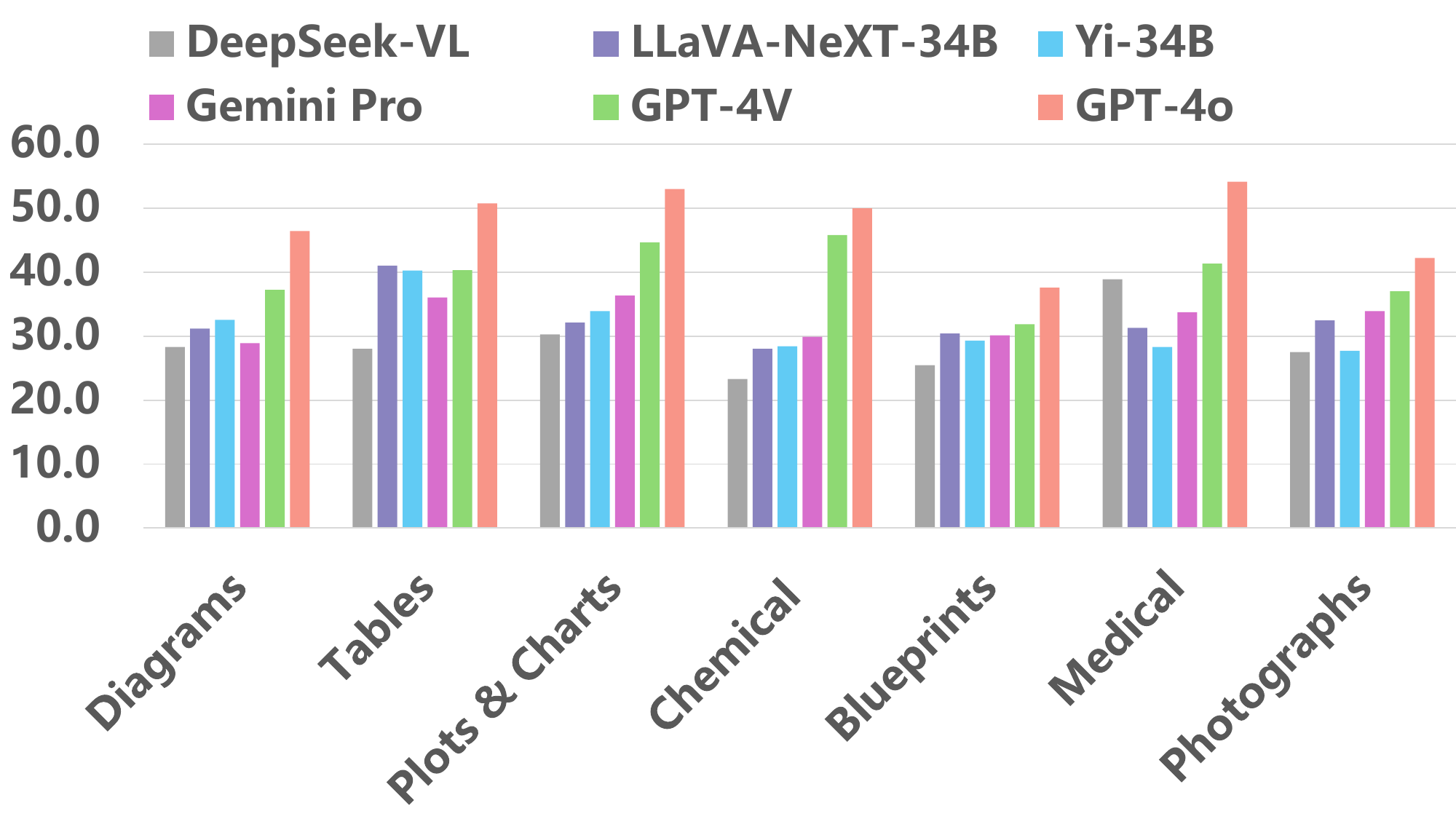}
      \includegraphics[width=0.48\linewidth]{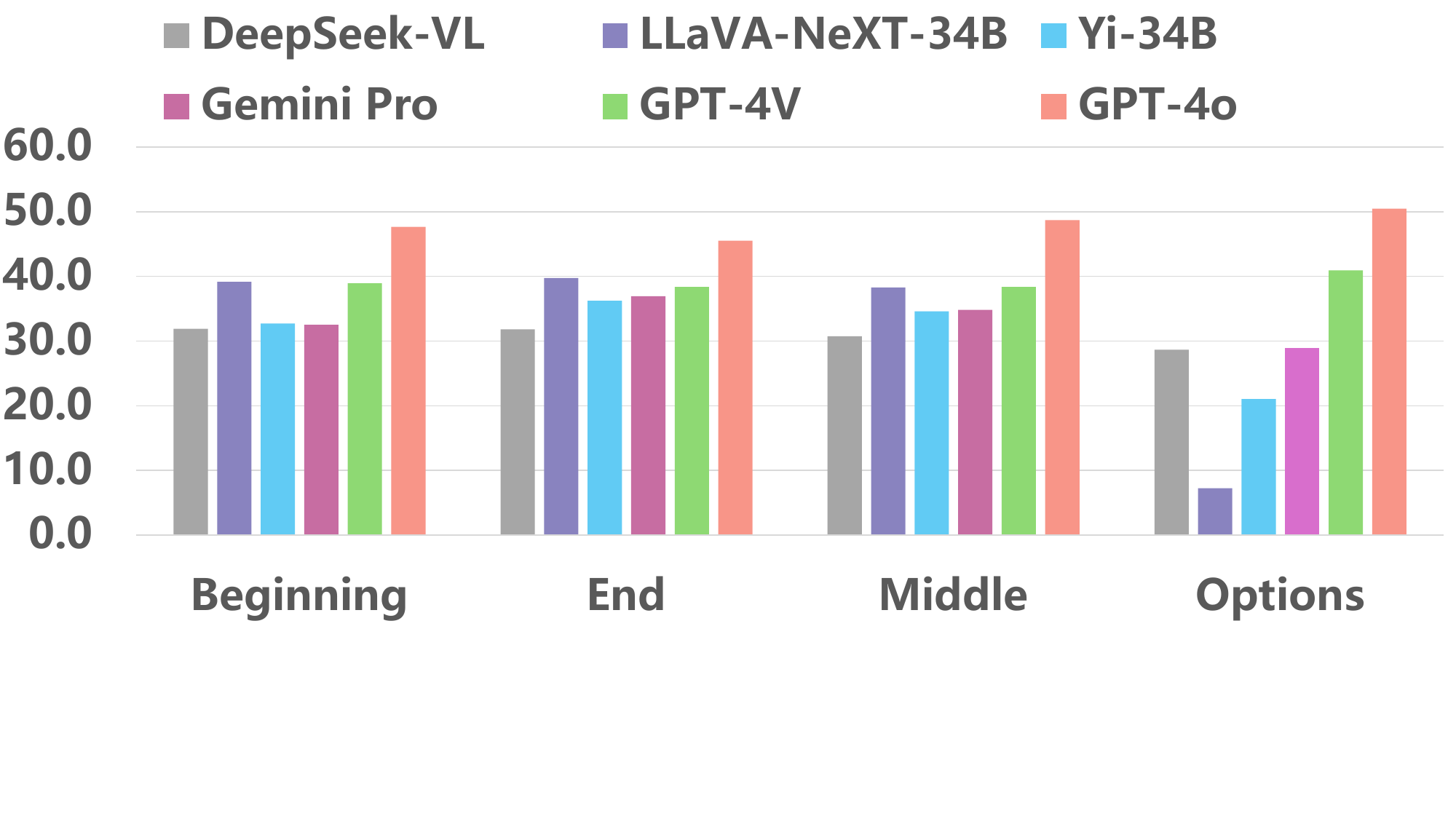}
   \caption{The zero-shot accuracy of different LMMs on different image types (Left) and positions (Right) on \our{} dataset.}
   \label{fig:img_type_pos}
\end{figure*}

\begin{table*}[t]
    \setlength{\tabcolsep}{10pt}
    \centering
    \small
    \caption{The detailed results of different LMMs on Health, Science and Engineering of \our{} dataset. Sci. and Eng. are short for Science and Engineering, respectively.}
    \begin{tabular}{l|ccc|ccc|ccc}
    \toprule
    \multirow{2}{*}{\bf Models} & \multicolumn{3}{c|}{\bf Chinese} & \multicolumn{3}{c|}{\bf English} & \multicolumn{3}{c}{\bf German} \\
    &  Health &  Sci. &  Eng. &  Health &  Sci. &  Eng.  &  Health &  Sci. &  Eng. \\
    \midrule
    Yi-VL-6B & 31.2 & \underline{34.1} & \underline{34.9} & 32.1 & 32.2  & 30.0 & 29.0 & 29.2 & 30.8 \\
    DeepSeek-VL & \bf 40.1 & 22.6 & 28.5 & \bf 38.0 & 31.9 & 28.6 & \underline{35.2} & 29.3 & 27.8  \\
    Yi-34B & 32.9 & \underline{34.1} & 33.6 & 34.0 & \underline{33.2} & \underline{32.6} & 29.4 & \underline{30.2} & \underline{32.0} \\
    LLaVA-NeXT-34B & \underline{38.1} & \bf 40.4 & \bf 37.0 & \underline{37.2} & \bf 36.8 & \bf 34.7 & \bf 36.9 & \bf 34.2 & \bf 34.5 \\
    \midrule
    Gemini 1.0 Pro &  38.8 & 34.5 & 31.4 & 34.9 & 33.4 & 29.8 & 33.1 & 30.5 & 28.8 \\
    $\quad$+ \textit{Chain-of-thought prompting} & 37.8 & 33.3 & 32.6 & 38.8 & 33.3 & 30.6 & 37.8 & 32.2 & 31.8 \\
    GPT-4V(ision) & 41.9 & 39.3 & 37.9 & 43.9 & 37.8 & 36.6 & 41.1 & 36.0 & 34.6 \\
    $\quad$+ \textit{Chain-of-thought prompting} & \underline{43.9} & \underline{46.2} &  \underline{41.7} & \underline{45.8} & \underline{43.3} & \underline{41.9} & \underline{42.5} & \underline{39.1} & \underline{39.3} \\
    GPT-4o & \bf 56.0 & \bf 47.3 & \bf 45.0 & \bf 56.2 & \bf 44.3 & \bf 42.8 & \bf 52.9 & \bf 40.9 & \bf 43.0 \\
    \bottomrule
    \end{tabular}
    \label{tab:subject}
\end{table*}

\begin{figure*}[t]
    \centering
    \includegraphics[width=0.85\linewidth]{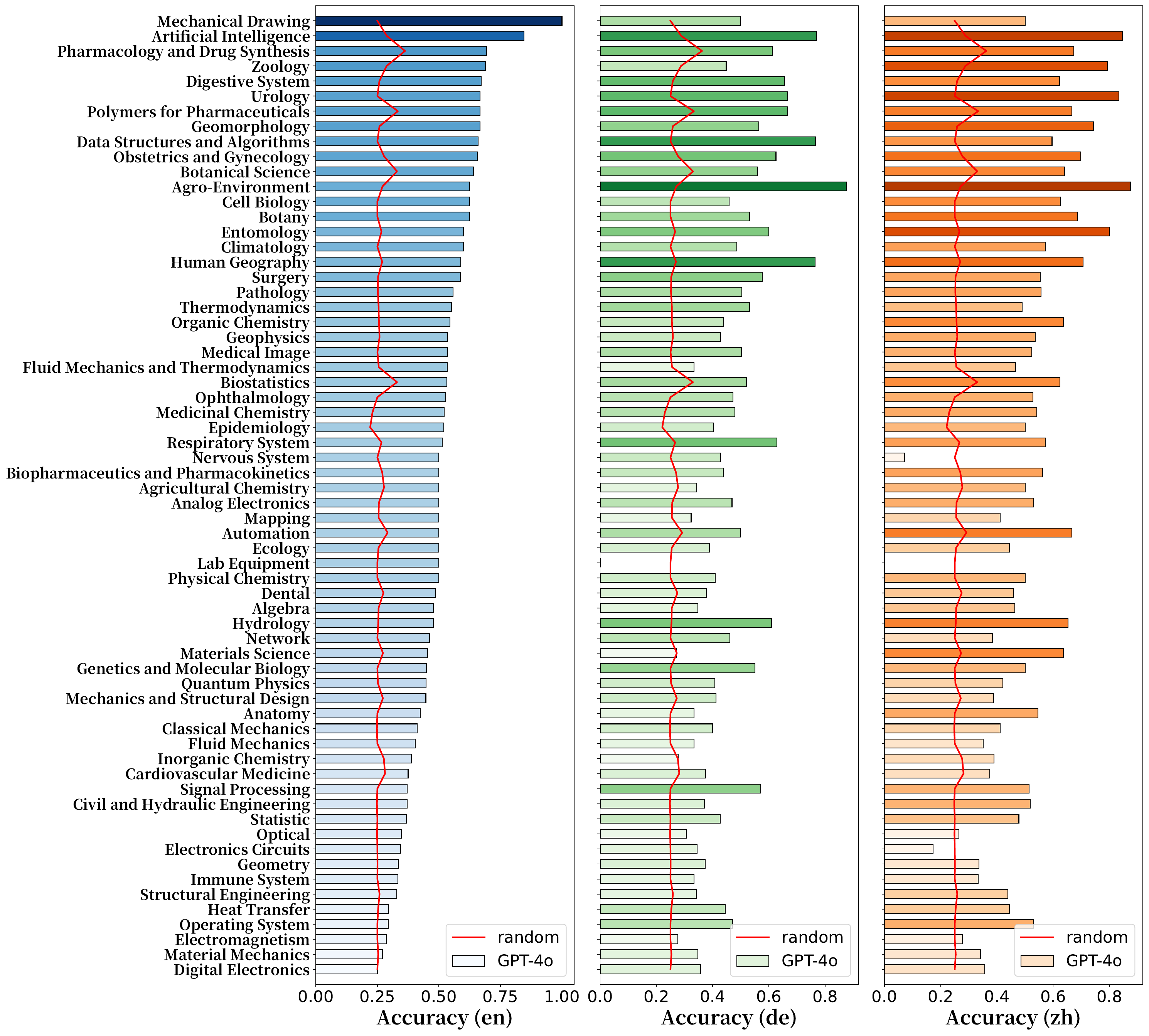}
    \caption{The zero-shot accuracy of GPT-4o across 64 subjects on M4U dataset.}
    \label{fig:subject_gpt4o}
\end{figure*}

\subsection{Models}

\textbf{LMMs.}
  For the open-sourced models, we select VisualGLM~\cite{visualglm}, Ying-VLM~\cite{yingvlm}, InstructBLIP series~\cite{instructblip}, InternLM-XComposer~\cite{internlmx}, CogVLM-Chat~\cite{cogvlm}, Qwen-VL-Chat~\cite{qwenvl}, Yi-VL-series~\cite{yi}, DeepSeek-VL~\cite{deepseekvl} and LLaVA-NeXT series~\cite{llavanext}. For closed source models, we choose Gemini 1.0 Pro~\cite{gemini}, GPT-4V(ision)~\cite{gpt4} and GPT-4o~\cite{gpt4o} using the provided API, \texttt{gemini-pro-vision}, \texttt{gpt-4-vision-preview} and \texttt{gpt-4o}, respectively. As for the augmented LMMs, we evaluate the performance of Gemini 1.0 Pro and GPT-4V with the chain-of-thought prompting.

\textbf{LLMs.}
We select Mistral-Instruct-v0.2-7B~\cite{mistral}, Qwen-1.5-7B-Chat, Qwen-1.5-14B-Chat~\cite{qwen} and Gemini 1.0 Pro (\texttt{gemini-pro}) for the open and closed source LLMs. We use Gemini 1.0 Pro (\texttt{gemini-pro-vision}) to generate the detailed caption in Chinese, English and German for each image. The visual captions are placed at the beginning of the prompt. The annotated image positions are used to refer each image.

\subsection{Main results}
\label{sec:main}

In this section, we present the comprehensive evaluation results of 22 leading LMMs and 4 LLMs with different prompt strategies. Table~\ref{tab:zs} demonstrates the performance of various LMMs and LLMs across Chinese, English and German on \our{} dataset. 

For the text-only LLMs, we first only use the text part of question to prompt these models. As shown in Table~\ref{tab:zs}, Qwen-1.5-14B Chat has only 25.1\% average accuracy on \our{} dataset, which is lower than 25.9\% of random choices. It proves that \our{} requires significant visual efforts to answer these questions. Further, we equip these LLMs with the detailed visual captions generated by Gemini 1.0 Pro. Qwen-1.5-14B Chat with additional captions outperforms itself without any visual information by a gain of 7.7\%, and achieves 32.8\% average accuracy, the highest scores among the baselines. Mistral-Instruct-v0.2-7B has 25.6\% average accuracy, since it does not follow the instruction to generate the valid option. We observe that Mistral-Instruct-v0.2-7B tends to reject to give an answer when not being provided with enough visual information. 

For LMMs, most of them do not have the satisfactory results on \our{} dataset. The state-of-the-art model, GPT-4o, achieves only 47.6\% average accuracy with zero-shot prompting. It indicates that \our{} is quite challenging for the existing models, and the reasoning capability of the multimodal models still has much room for future improvement. With the powerful LLM, Nous-Hermes Yi-34B~\footnote{\url{https://huggingface.co/NousResearch/Nous-Hermes-2-Yi-34B}}, LLaVA-NeXT-34B scores highest among the open-source LMMs, even significantly outperforms Gemini 1.0 Pro by a gain of 3.8\% on average accuracy. As for augmented LMMs, chain-of-thought prompting further boosts the performance. GPT-4V with chain-of-thought prompting outperforms itself with zero-shot prompting by a gain of 3.8\% on average accuracy. It demonstrates that explicitly generating the reasoning steps is also beneficial for complex multimodal reasoning.

Furthermore, we observe that the existing models has strong language preferences on multilingual multimodal reasoning tasks. InstructBLIP Vicuna-7B achieves 28.1\% accuracy on the English part of \our{}, while only has 13.7\% and 19.7\% accuracy on the Chinese and German part, respectively. For GPT-4V, the average accuracy on the Chinese and English is both 3\% higher than on the German. Besides, we observe that the effect of chain-of-thought prompting also differs across different languages. For instance, chain-of-thought improves the performance of Gemini 1.0 Pro on English and German part by a gain of 1.5\% and 3.1\% accuracy, while leads to a degradation of 0.5\% on Chinese part. We argue that this results from the lack of the multilingual vision-language corpus used for multimodal training, and the LLMs of these LMMs (e.g., Vicuna-7B, Vicuna-13B) do not well support the multilingual capability. 

\subsection{Cross-lingual Multimodal Evaluation} 
\label{sec:cross_lingual}

The cross-lingual multimodal questions refer to the samples whose images contains the key concept or label in one language, while the text instruction is in another language. These types of questions have wide applications in real life. A typical scenario is when an English-native speaker queries the possible uses of an item from a Chinese instruction manual. We investigate whether the model can not only perceive textual content in multiple languages, but also reason based on multilingual information from visual inputs.

A common approach is to first translate the question into the language of the visual content, turning it into a monolingual problem. We compared the end-to-end method to the two-stage translate-then-questioning. For the end-to-end method, we directly require the model to generate a short option letter given the original question and images. As for the translate-then-questioning methods, since the images may contain the key concept or label in Chinese, we first use the model to translate the question part into Chinese. Then we require the model to provide a short option letter given the translated questions and images.

\begin{table}[t]
    \centering
    \small
    \setlength{\tabcolsep}{8pt}
    \caption{Comparison of the two-stage translate-then-questioning and end-to-end method on the cross-lingual set of M4U-mini.}
    \begin{tabular}{lcccccc}
        \toprule
       \bf Methods & \bf EN & \bf DE & \bf AR & \bf JA & \bf TH & \bf Avg. \\
        \midrule
        Two-stage & \bf 41.1 & 41.7 & 26.9 & 33.1 & 31.6 & 34.9 \\
        \bf End-to-end & 35.4 & \bf 42.8 & \bf 38.4 & \bf 39.0 & \bf 40.1 & \bf 39.1 \\
        \bottomrule
    \end{tabular}
    \label{tab:two_stage}
\end{table}

\begin{figure}[t]
    \centering
    \includegraphics[width=\linewidth]{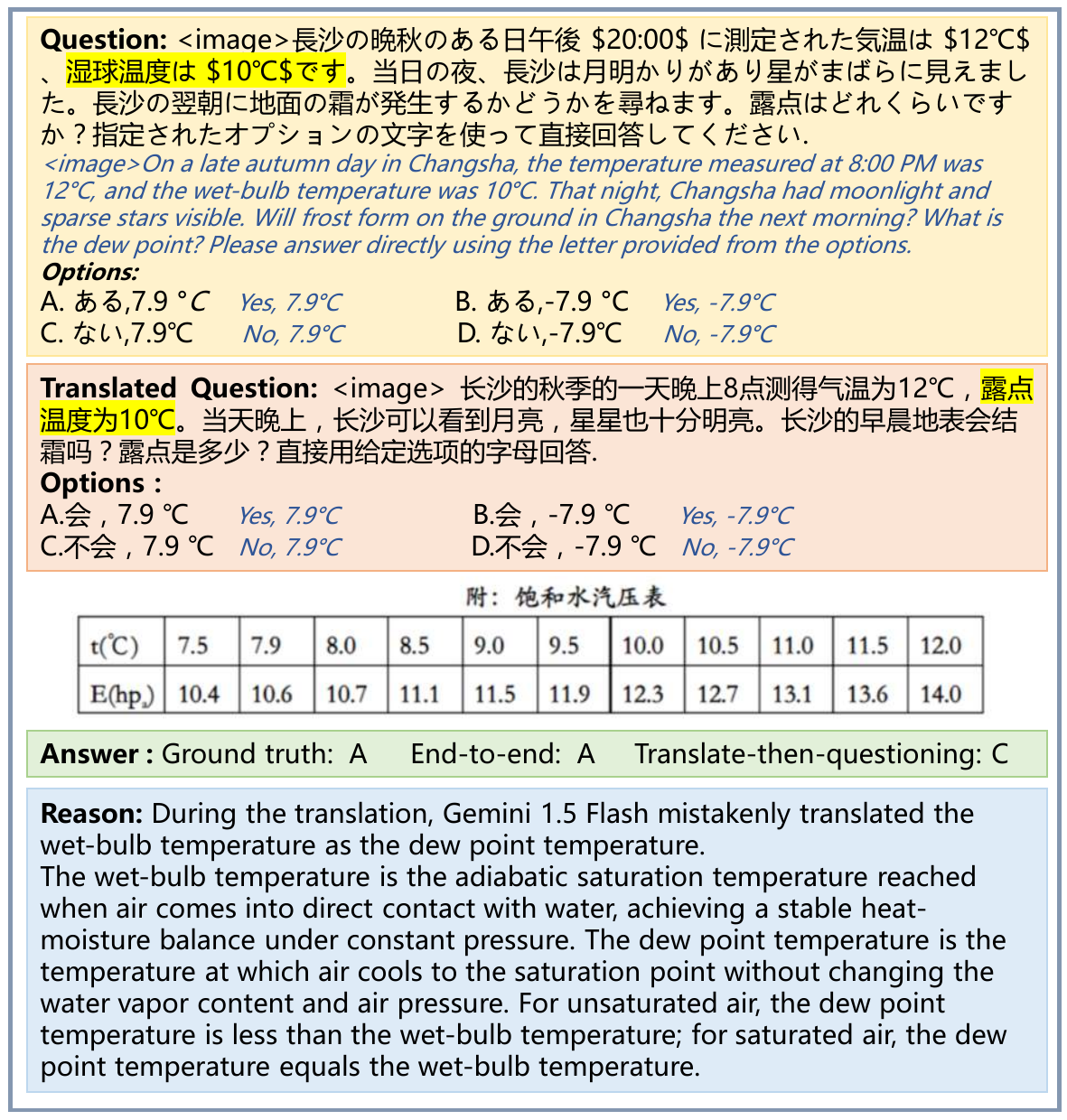}
    \caption{A typical failed case of two-stage translate-then-questioning method.}
    \label{fig:two_stage_error}
\end{figure}

We found that the two-stage approach often introduces hallucinations and errors during translation stage, especially in the medium- or low- resources languages that the model struggles with.  As shown in the Table~\ref{tab:two_stage}, we measured the difference between these two methods with Gemini 1.5 Flash on the cross-lingual set of M4U-mini where the content of the images is in Chinese, and the questions are multilingual. We translate the question part in the other languages into Chinese, then perform visual question answering. The results show that the end-to-end approach outperforms the translation-then-questioning method by a gain of 4.2\% on average scores. A similar phenomenon was also observed in M3Exam~\cite{m3exam}: directly translating other languages into English for questioning often led to a decline in performance. This is because some questions are closely tied to specific languages, and direct translation often loses this context, such as with country-specific information. We provide a typical failed case of translation-then-questioning in Figure~\ref{fig:two_stage_error}. During translation, Gemini 1.5 Flash incorrectly rendered "wet-bulb temperature" as "dew point temperature," resulting in a wrong answer.

To measure the cross-lingual multimodal capability of the LMMs, we select a subset from \our{}: the image of each sample in this subset contains the text that labels or explains the key concepts or objects in the picture, while the textual description of the question is multilingual. For example, as illustrated in Figure~\ref{fig:samples}, the visual content contains the key text in Chinese that labels bond length between atoms and explains the single-layer structure of the material, and the question part is multilingual. The models are required to perform complex reasoning given the multilingual both textual and visual contents. The cross-lingual set contains 1065, 417 and 531 samples from Science, Engineering and Healthcare, respectively, resulting in up to 2013 samples in total. 

We evaluate the performance of different LMMs on the cross-lingual set of \our{}. As shown in Table~\ref{tab:cross_lingual}, almost all models suffer from a degradation of performance when the image contains the key textual information in Chinese but the question is English or German. It shows that these models are short for following multilingual instructions to understand the visual contents with the textual information of another language. Furthermore, as for the augmented LMMs, we observe that the chain-of-thought prompting significantly improves the performance of GPT-4V on English and German. This is aligned with our previous evaluations on the full set of \our{} in Table~\ref{tab:zs}.

\begin{table*}[t]
    \setlength{\tabcolsep}{12pt}
    \centering
    \small
    \caption{The zero-shot accuracy of various LLMs and LMMs on the \our{}-mini dataset.}
    \label{tab:m4u-mini}
    \begin{tabular}{l|ccccccc}
    \toprule
    \bf Models &  \bf English & \bf Chinese & \bf German & \bf Japanese & \bf Arabic & \bf Thai & \bf Avg.\\
    \midrule
    Random choice & \multicolumn{7}{c}{26.3} \\
    \midrule
    \multicolumn{6}{l}{\ \emph{Large Language Models}} \\
    Qwen1.5-14B-Chat~\cite{qwen} & 28.9 & 17.7 & 29.5 & 19.3 & 26.9 & 12.0 & 22.4 \\
    Qwen2.5-14B-Instruct~\cite{qwen2_5} & 35.0 & 25.7 & 25.7 & 13.6 & 35.7 & 13.8 & 24.9 \\ 
    \midrule
    \multicolumn{6}{l}{\ \emph{Large Multimodal Models}} \\
    DeepSeek-VL-Chat~\cite{deepseekvl} & 35.4	& 33.6	& 35.0	& 32.1	& 24.8	& 25.4	& 31.0 \\
    Phi-3.5-Vision-Instruct~\cite{phi3} & 34.3 & 27.2 & 33.4 & 30.4 & 31.7 & 30.9 & 31.3 \\
    LLaVA-NeXT-34B~\cite{llavanext} & 44.1 & 44.2 & 39.0	& 36.0 & 11.4 & 34.0 & 34.8 \\
    InternVL-2.5-8B~\cite{internvl2_5} & 41.7 & 38.5 & 38.3 & 36.1 & 31.4 & 31.7 & 36.3 \\
    Gemini 1.5 Flash~\cite{gemini1_5} & 35.4	& 46.3	& 42.8	& 39.0	& 38.4	& 40.1 & 40.3\\
    Qwen2-VL-7B-Instruct~\cite{qwen2vl} & 43.5 & 46.6 & \underline{44.1} & \underline{47.6} & \underline{41.5} & \underline{41.4} & 44.1 \\
    InternVL-2.5-26B~\cite{internvl2_5} & \underline{44.2} & \underline{51.3} & \bf 48.1 & 46.4 & 37.6 & 37.3 & \underline{44.2} \\
    GPT-4o~\cite{gpt4o} & \bf 44.9 & \bf 53.7 & 42.4	& \bf 49.1 & \bf 45.2 & \bf 48.8 & \bf 47.3\\
    \bottomrule
    \end{tabular}
\end{table*}

\subsection{Fine-grained results}
\textbf{Different Image Types and Positions.} We demonstrate the visualization of the detailed results of various LMMs on different image types and positions in Figure~\ref{fig:img_type_pos}. We reclassified 13 image types into 7 categories based on the style and application of the image. As shown in the left part of Figure~\ref{fig:img_type_pos}, GPT-4o shows impressive performance on the image type of Plots \& Charts and Medical compared with the other models, but has unsatisfactory results on Blueprints. We argue that this is because the Blueprints contain many engineering sketches that require the capability of the fine-grained perception and domain-specific knowledge about engineering standards. \our{} not only supports the image-text pairs as the input, but includes interleaved image-text documents. Thus, we conduct the analysis about the performance of the selected LMMs on different positions of the images. We divide these questions into four groups according to the image position: image at the beginning, end, middle of the question and in the options. As shown in the right part of Figure~\ref{fig:img_type_pos}, on the questions with images in the options, GPT-4o and GPT-4V outperform the other models by a large gain, and LLaVA-NeXT-34B performs poorly on this types of the questions. We argue that this is because the LLaVA-NeXT series are only trained with a high-quality corpus of image-text pairs. Instead DeepSeek-VL is pre-trained with a large mixture of image-text pairs and interleaved documents, and it does not suffer from a significant degradation of performance on the questions with images in the options.

\textbf{Different Subjects, Disciplines and Languages.} We present the detailed results of various LMMs on different fields of Chinese, English and German in Table~\ref{tab:subject}. GPT-4o outperforms the other models by large improvements on all fields of all languages. For the open-source models, we observe that LLaVA-NeXT-34B shows impressive results on scientific reasoning, and DeepSeek-VL demonstrates good performance on Health. Further, we observe that on Science, the chain-of-thought prompting significantly improves the performance of GPT-4V by a gain of over 6\% accuracy in Chinese and English, while only boosts the performance by an improvement of 3.1\% accuracy on German. The similar phenomenon also exists for Gemini 1.0 Pro. On Health part, Gemini 1.0 Pro with the chain-of-thought prompting outperforms it with zero-shot prompting by a gain of 4.9\% and 4.7\% on English and German, but it leads to a degradation of 1.0\% accuracy on Chinese. These results show that the effect of the chain-of-thought prompting also differ from different languages. Furthermore, we visualize the detailed results of GPT-4o across 64 subjects in Figure~\ref{fig:subject_gpt4o} on the test set of \our{}.

\section{Evaluation on \our{}-mini}

We select the state-of-the-art open-source LLMs, Qwen-1.5-14B-Chat~\cite{qwen} and Qwen2.5-14B-Instruct~\cite{qwen2}, as reference baselines to assess the visual challenges posed by \our{}-mini. For multimodal models, we include LLaVA-NeXT-34B~\cite{llavanext}, DeepSeek-VL-Chat~\cite{deepseekvl}, Phi-3.5-Vision~\cite{phi3}, Qwen2-VL-7B~\cite{qwen2vl}, and the InternVL2.5 series~\cite{internvl2_5}. For closed-source models, we evaluate GPT-4o~\cite{gpt4o} and Gemini-1.5-Flash~\cite{gemini1_5} using their respective APIs.

We evaluate the zero-shot performance of these models on \our{}-mini, and Table~\ref{tab:m4u-mini} summarizes the results across six languages. Without any visual information, Qwen-1.5-14B-Chat and Qwen2.5-14B-Instruct achieve only 22.4\% and 24.9\% average accuracy, respectively—both lower than the 26.3\% baseline of random choice. This demonstrates that \our{}-mini relies heavily on visual understanding and contains content less exposed in the training data of standard LLMs.

Furthermore, we observe that GPT-4o consistently achieves strong performance across all languages. In contrast, open-source models, while competitive in certain languages—such as InternVL-2.5-26B and Qwen2-VL-7B-Instruct, which both achieve around 44\% average accuracy—generally lag behind the closed-source models. Notably, LLaVA-NeXT-34B experiences significant performance drops in medium- and low-resourced languages (e.g., scoring only 11.4\% in Arabic), highlighting potential limitations in its training data coverage.

\begin{figure*}[t]
    \centering
      \centering
      \includegraphics[width=0.75\linewidth]{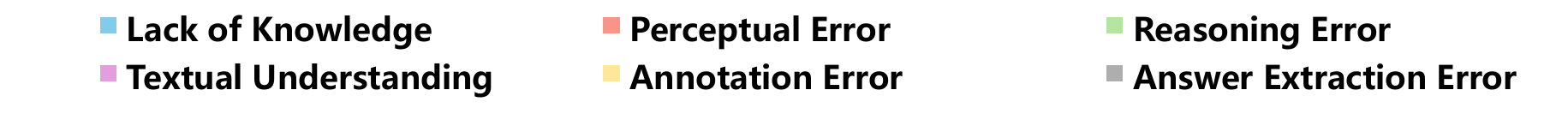}
      \includegraphics[width=0.25\linewidth]{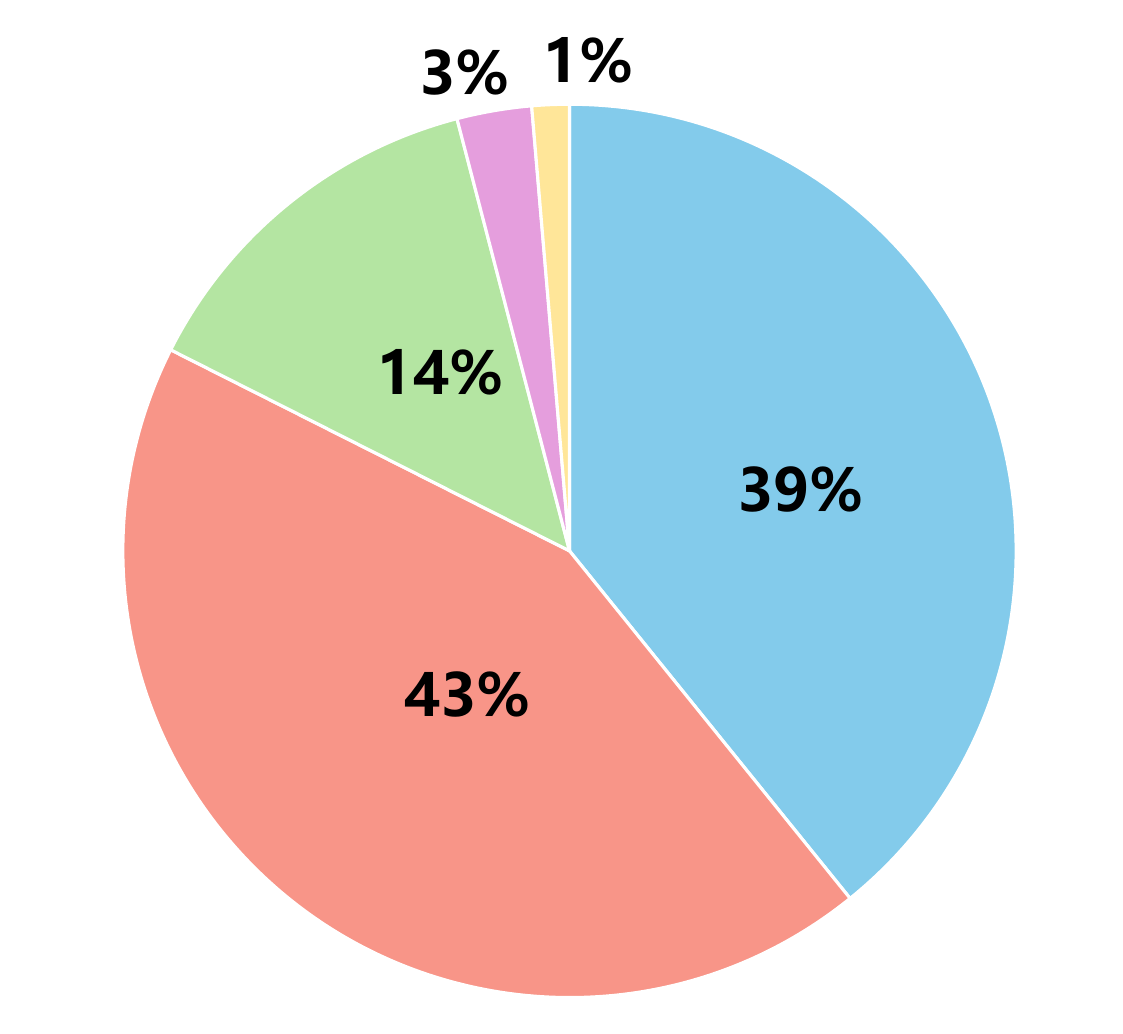}
      \includegraphics[width=0.25\linewidth]{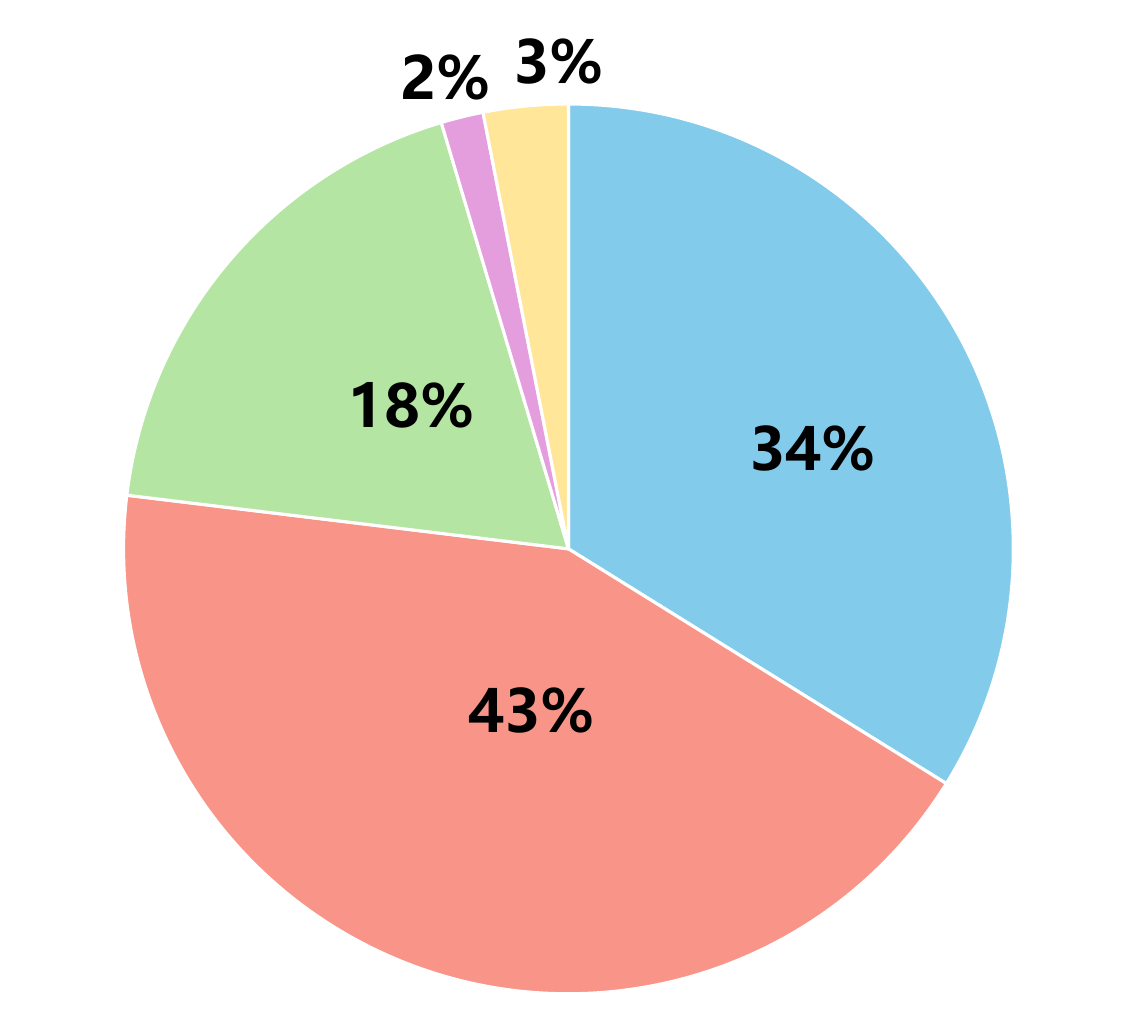}
      \includegraphics[width=0.25\linewidth]{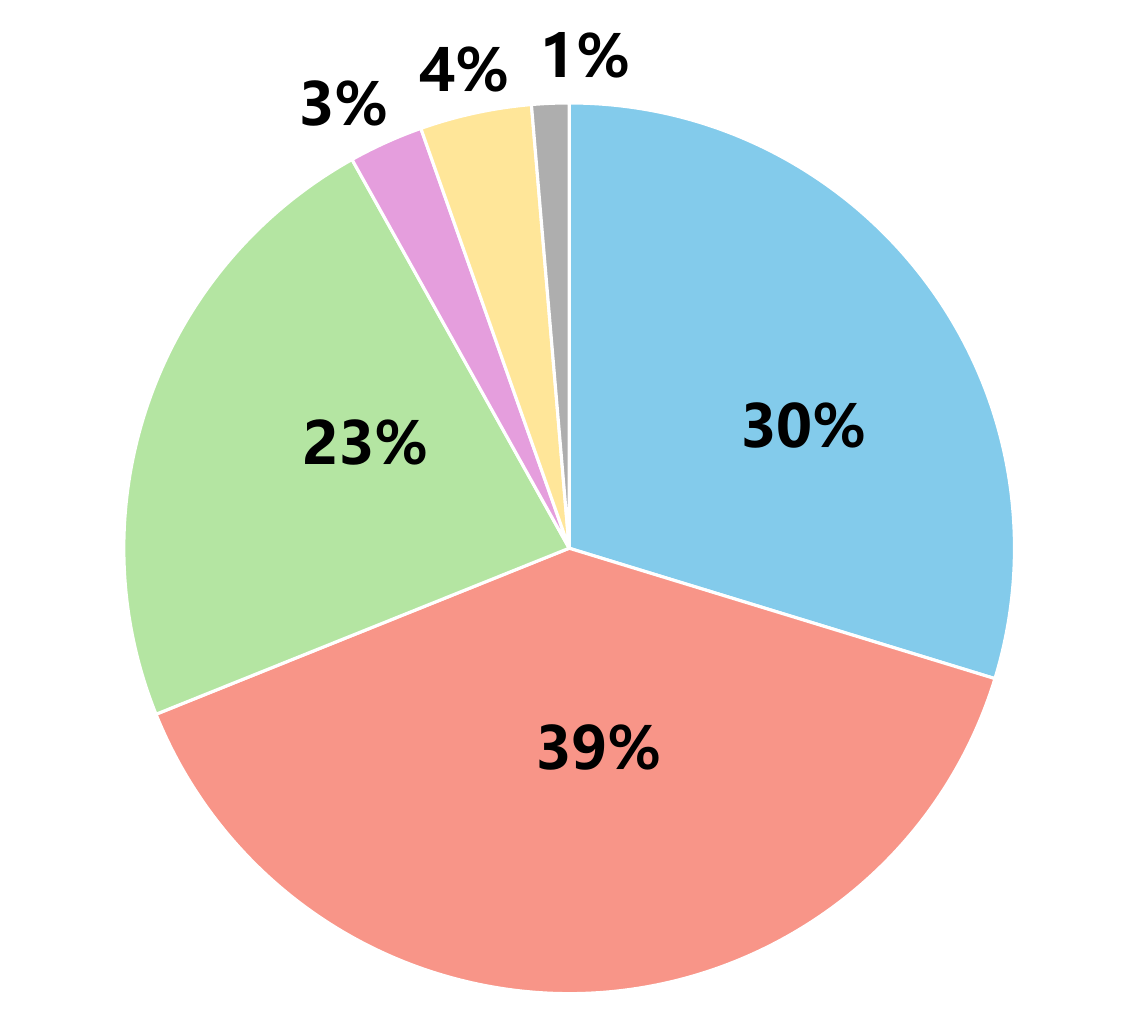}
   \caption{The distribution of the wrong cases of GPT-4V in different languages (Left: Chinese, Middle: English, Right: German).}
   \label{fig:error}
\end{figure*}

\begin{figure}[t]
    \centering
    \includegraphics[width=0.49\textwidth]{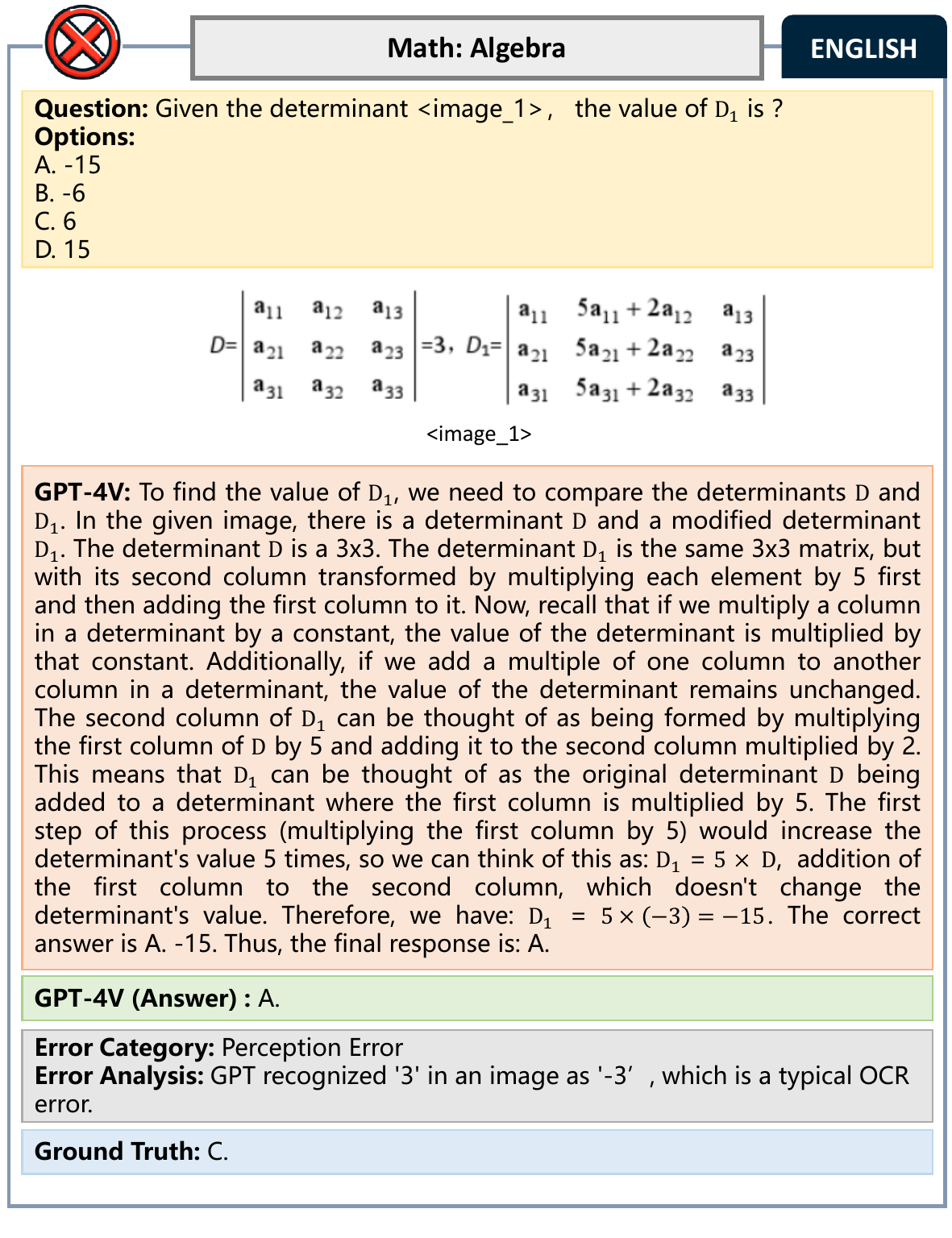}
    \caption{A typical case of perceptual error (subject: math, language: English).}
    \label{fig:case1_en}
\end{figure}

\begin{figure}[t]
    \centering
    \includegraphics[width=0.5\textwidth]{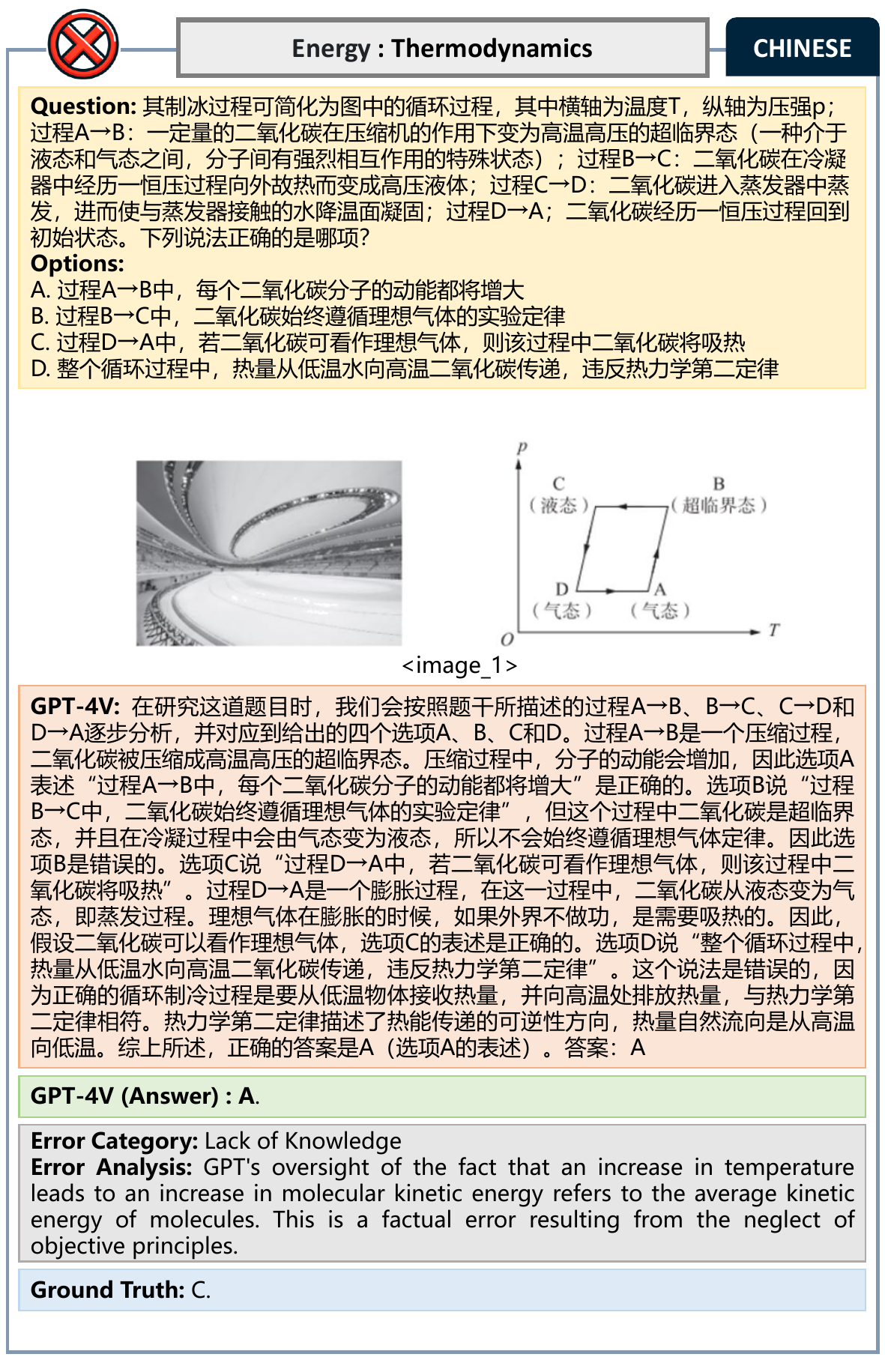}
    \caption{A typical case of lack of knowledge (subject: energy, language: Chinese).}
    \label{fig:case5_cn}
\end{figure}

\begin{figure}[t]
    \centering
    \includegraphics[width=0.49\textwidth]{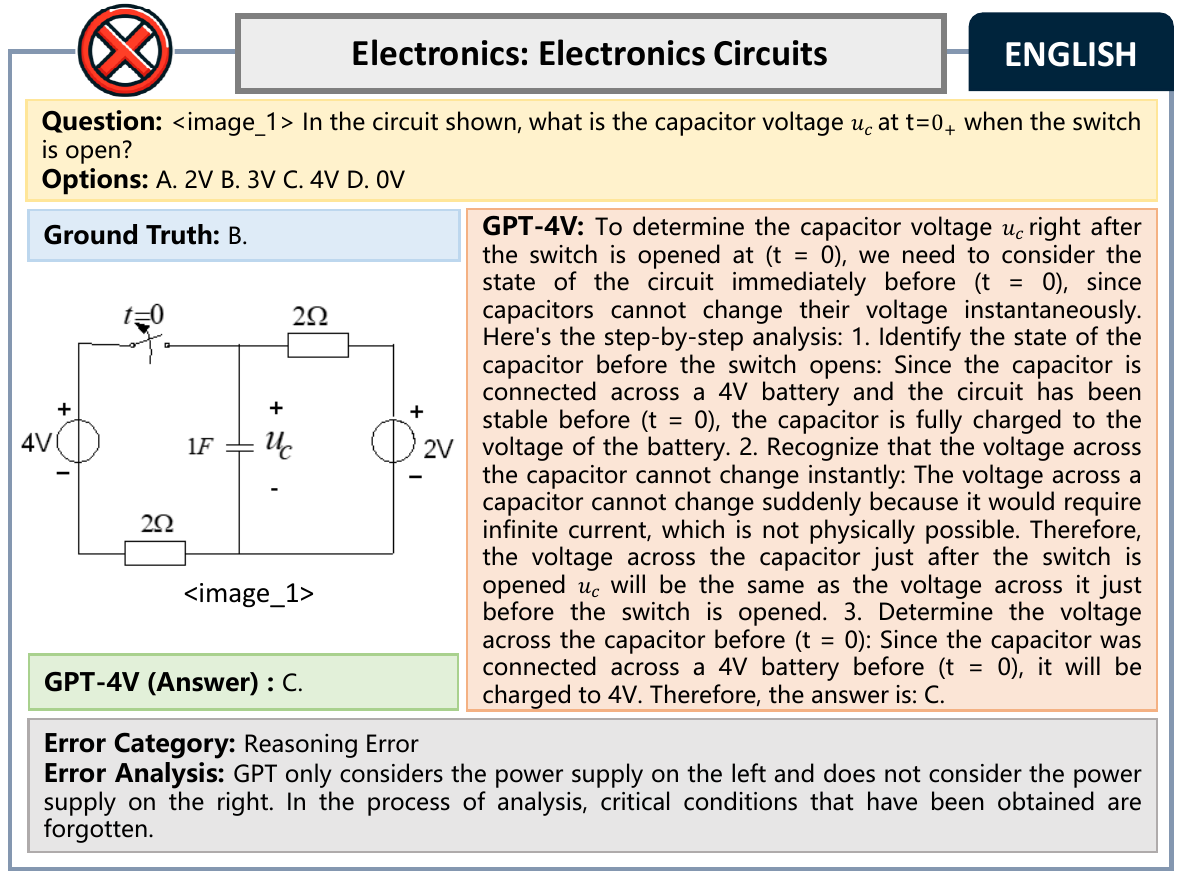}
    \caption{A typical case of reasoning error (subject: electronics, language: English).}
    \label{fig:case7_en}
\end{figure}

\section{Qualitative Analysis}
\label{sec:case_study}

We conduct qualitative analysis for GPT-4V with the chain-of-thought prompting. Specifically, we randomly sample 75 questions (2.5\%) from different disciplines of each language. In these instances, GPT-4V has errors in responses and analysis in at least one language. We analyze the cause of these wrong cases, and divided them into six categories: perceptual error, lack of knowledge, reasoning error, textual understanding, annotation error and answer extraction error. 

The distribution of selected samples across different categories are illustrated in Figure~\ref{fig:error}. Perceptual error, lack of knowledge, and reasoning error account for the major causes of failed cases (96\% in Chinese, 95\% in English, and 92\% in German). We observe that GPT-4V tends to exhibit lack of knowledge on the Chinese part of \our{}, while reasoning errors are more likely to occur in German and English. We present more qualitative results in supplementary materiel.

\paragraph{Perceptual Error} Perceptual error is the most frequent error made by GPT-4V. It corresponds to the illusion phenomenon that occurs when extracting visual information from images provided by the questions. These kinds of hallucination could be divided in two main categories: visual information deficiency and misinterpretation.  
Figure~\ref{fig:case1_en} shows a typical case for the visual information misinterpretation: the extracted information is complete but contains mistakes. A portion of these mistakes are common perceptual errors in OCR and visual localization. 

\paragraph{Lack of Knowledge} We define the lack of knowledge as the model has factual misunderstanding about the key concepts in questions and provides erroneous premise to the reasoning process. 
As shown in Figure~\ref{fig:case5_cn}, GPT-4V equates the average kinetic energy of a molecule to the kinetic energy of a single molecule, overlooking key preconditions of physical laws. 

\paragraph{Reasoning Error} The reasoning error is categorized as the mathematical miscalculations and logical errors in the analysis procedure, which often occur in subjects need numerical computations and logical inference, such as math, physics, and electronics. As demonstrated in Figure~\ref{fig:case7_en}, GPT-4V only considers the power supply on the left and does not consider the power supply on the right.

\paragraph{Others} Remaining error cases only occupy a small portion in selected cases, while depict long-tailed but various error reasons including textual misunderstanding, annotation error, and answer extraction error. Annotation error caused by typo or translation issues maintains less than $5\%$ after manually checked by annotators. 

\section{Conclusion}
\label{sec:future}

In this work, we introduce \our{}, a novel and challenging benchmark for evaluating the capability of multilingual multimodal understanding and reasoning. \our{} contains 10,005 multiple-choice questions, covering 64 disciplines across 16 subfields in Science, Engineering, and Healthcare in Chinese, English, and German. Table~\ref{tab:benchmark} demonstrates that \our{} requires significant visual efforts compared with M3Exam. As shown in Table~\ref{tab:zs}, the state-of-the-art model, GPT-4o, achieves only 47.6\% average accuracy with zero-shot prompting, indicating that \our{} is quite challenging for existing models. Furthermore, we observe that the leading LMMs exhibit significant language preferences. These results demonstrate that there is still significant room for improvement in LMMs, especially in expert-level multilingual multimodal reasoning.

Currently \our{} focuses on the evaluation of science problems for multimodal reasoning. In the future, we aim to investigate the performance of multilingual LMMs on questions associated with cultural backgrounds (e.g., history and politics). Additionally, we plan to include multilingual rationales for \our{} to construct a fine-grained metric that considers the correctness of both reasoning steps and final predictions. 

\section{Acknowledgments}

This work is partially supported by Natural Science Foundation of China under contract No. U21B2025, and National Key R\&D Program of China No. 2021ZD0111901, 2023YFF1105104. Hongyu Wang, Jiayu Xu and Senwei Xie contributed equally to this work. Hongyu Wang led the project and conducted the evaluation on open-sourced models. Jiayu Xu led the annotation team and was primarily responsible for data construction and quality control. Senwei Xie was primarily responsible for the evaluation of closed-source models and conducted qualitative analysis.

\bibliography{sample}
\bibliographystyle{IEEEtran}

\section*{Dataset Documentation}
\label{ap:doc}

\subsection{Usage}
\begin{itemize}
    \item \our{} and \our{}-mini dataset are public available at \url{https://huggingface.co/datasets/M4U-Benchmark/M4U} and \url{https://huggingface.co/datasets/M4U-Benchmark/M4U-mini}, respectively. We provide the evaluation code for LLaVA and GPT-4o at \url{https://github.com/M4U-Benchmark/M4U}.
    \item The dataset is saved in Parquet format. We present the data format and the examples in the README.md file. Besides, we also provide the example codes to show how to evaluate the multimodal models on the dataset.
    \item License: \our{} is under CC BY-NC-SA Liences. The guidelines for annotators stress the importance of strictly following copyright and licensing rules from the original data sources, particularly avoiding materials from websites that prohibit copying and redistribution. If you come across any data samples that may violate copyright or licensing regulations, please inform us. Once verified, such samples will be promptly removed.
\end{itemize}

\subsection{Data sources}

\our{} consists of 10,005 multiple-choices questions, covering 64 disciplines of 16 subfields from Science, Engineering and Healthcare. To minimize the risk of data contamination, the samples are collected from college exams, the quizzes of online video lectures. Further a large portion (35\%) of the questions are written by our team according to the textbooks. 
The guidelines for annotators stress the importance of strictly following copyright and licensing rules from the original data sources, particularly avoiding materials from websites that prohibit copying and redistribution. If you come across any data samples that may violate copyright or licensing regulations, please inform us. Once verified, such samples will be promptly removed.

\subsection{First-stage Processing Guidelines}
\label{ap:principle}
We summarize the detailed first-stage processing guidelines for the annotators. For each dimension, we require the annotator to score the sample following the below guidelines. The question with the higher scores indicates higher quality. We filter out the questions with average scores lower than 2.0. As for the image quality, the standard is:
\begin{itemize}
    \item 0 score: The image is extremely blurry, difficult to recognize, or most of it is cropped, resulting in severe information loss.
    \item 1 score: The image is relatively blurry, details are hard to discern, or parts of the image are cropped, leading to some information loss.
    \item 2 score:  The image is slightly blurry; most content is recognizable but details are unclear, or the image is slightly cropped, but most information is complete.
    \item 3 score: The image is mostly clear; all major content is recognizable, though some details may not be clear.
    \item 4 score: The image is clear; all content and details are easily recognizable with no apparent defects.
    \item 5 score: The image is very clear; details are excellently represented, complete without any cropping or obstructions, meeting or exceeding the expected quality standards.
\end{itemize}
The guideline for measuring the question description quality is:
\begin{itemize}
    \item 0 score: The question is vague and completely unintelligible, with no clear intent.
    \item 1 score: The question statement is ambiguous, difficult to fully understand its intent, with multiple possible interpretations.
    \item 2 score: The question statement is basically clear, but there are some ambiguities or lack of rigor that need further clarification.
    \item 3 score: The question statement is clear, though there are some details that are not rigorous or there is slight ambiguity.
    \item 4 score: The question statement is both clear and rigorous, with details well handled, and only very minor issues present.
    \item 5 score: The question statement is extremely clear and rigorous, logical, without any ambiguity, fully meeting high standards.
\end{itemize}
The standard for measuring the difficulty of visual understanding is:
\begin{itemize}
    \item 0 score: The question almost does not rely on visual ability, can be fully understood without any visual information.
    \item 1 score: The question does not completely rely on visual ability, both visual and non-visual information are balanced.
    \item 2 score: Although the question relies on visual ability, a considerable proportion of non-visual information assists understanding.
    \item 3 score: The question largely depends on visual ability, but some non-visual information is provided.
    \item 4 score: The question greatly depends on visual ability, with very little content provided by non-visual information.
    \item 5 score: The question completely depends on visual ability, without it, the content is incomprehensible.
\end{itemize}

\begin{figure}[h]
    \centering
    \includegraphics[width=0.8\linewidth]{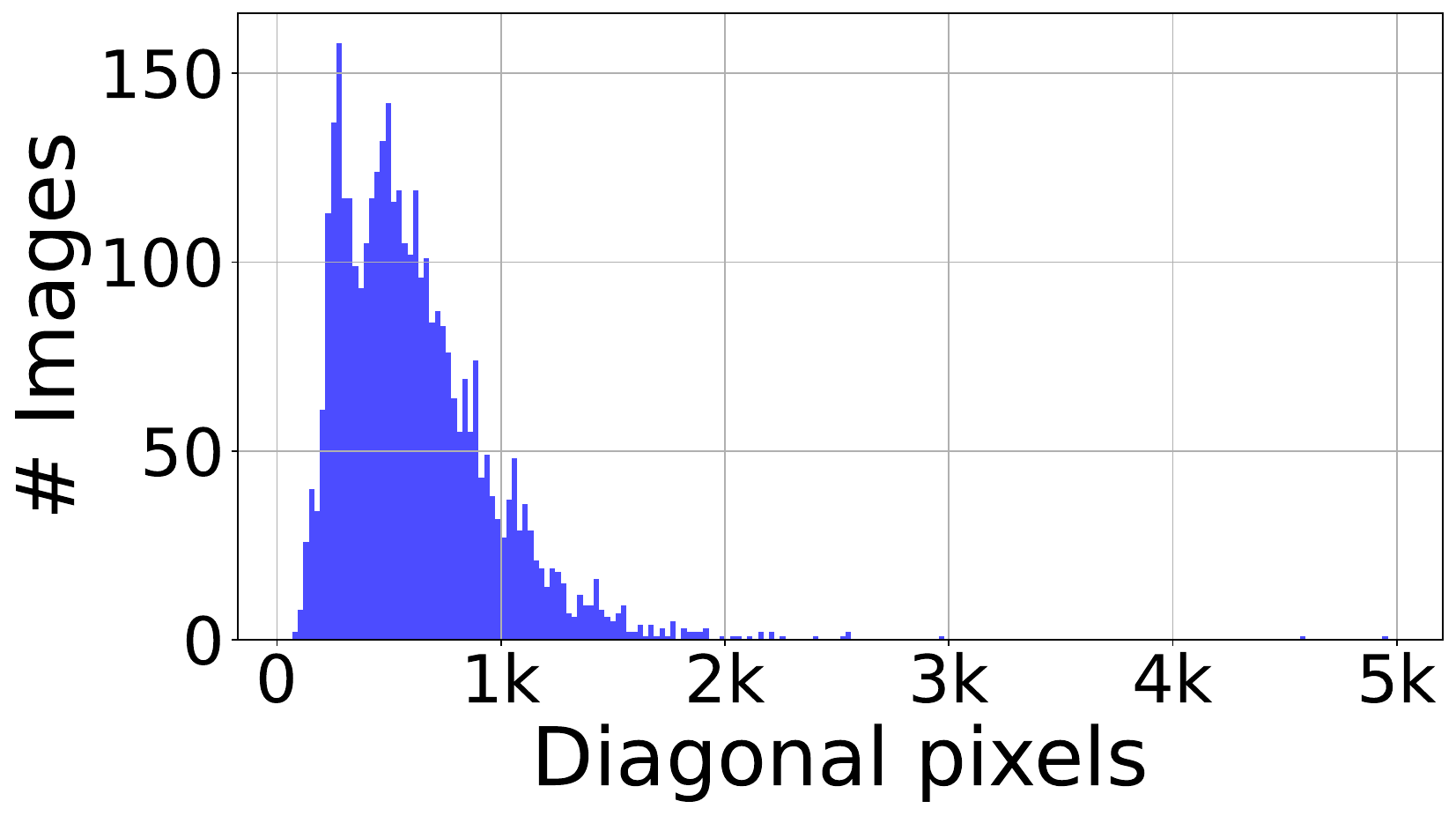}
    \caption{The distribution of image resolution for M4U dataset.}
    \label{fig:res}
\end{figure}

Furthermore, we recruit a team of graduate student of related majors to access the difficulty and correctness for the questions. The team will filter out the questions with wrong answer or minor visual efforts. We present the distribution of image resolution for M4U dataset in Figure~\ref{fig:res}.

\section*{Detailed Results of Qualitative Analysis}
\label{ap:case_study}

\begin{figure}[h]
    \centering
    \includegraphics[width=0.5\textwidth]{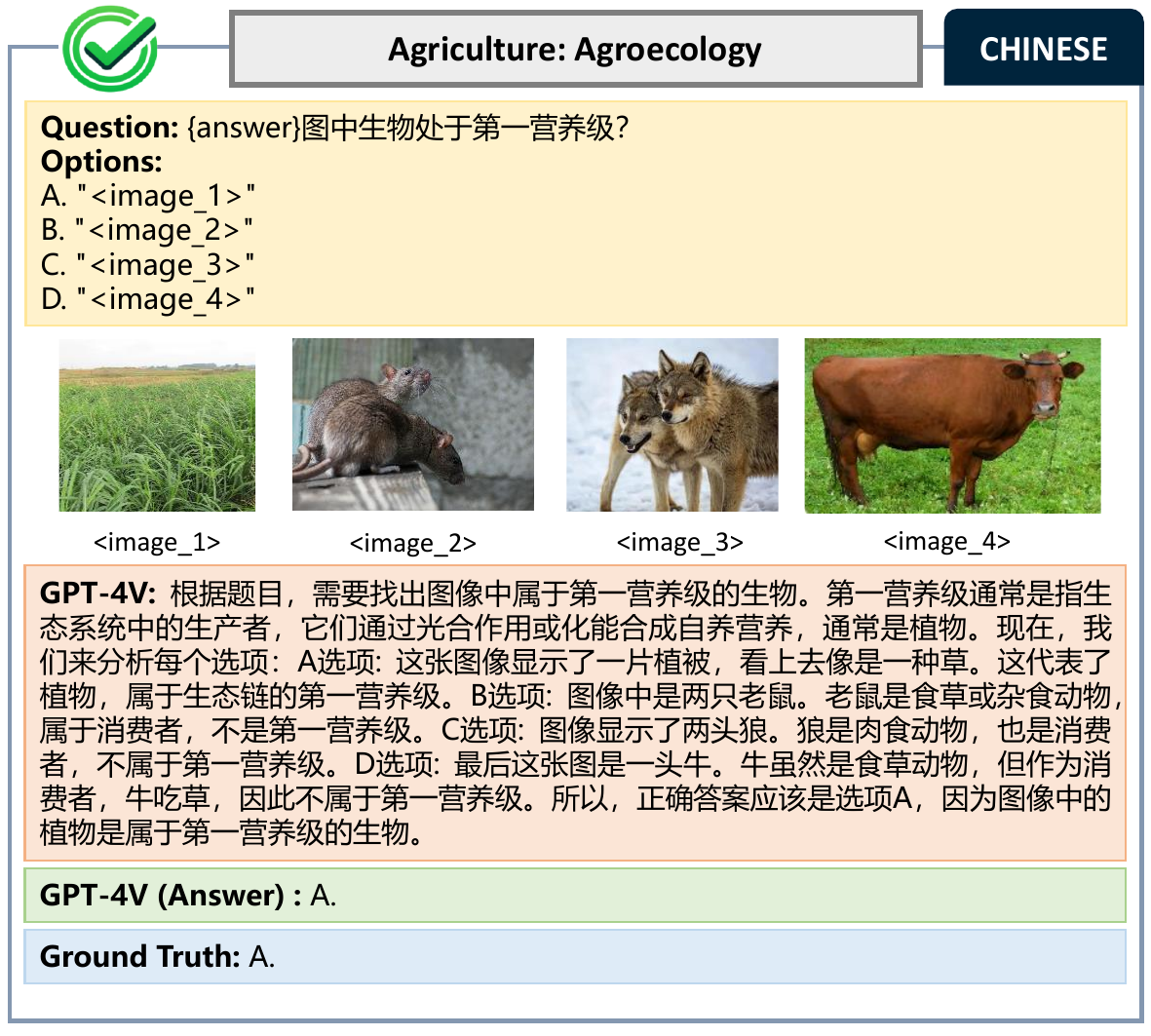}
    \caption{A sample correct case (subject: agriculture, language: Chinese).}
    \label{fig:case_true_cn}
\end{figure}
\begin{figure}[h]
    \centering
    \includegraphics[width=0.5\textwidth]{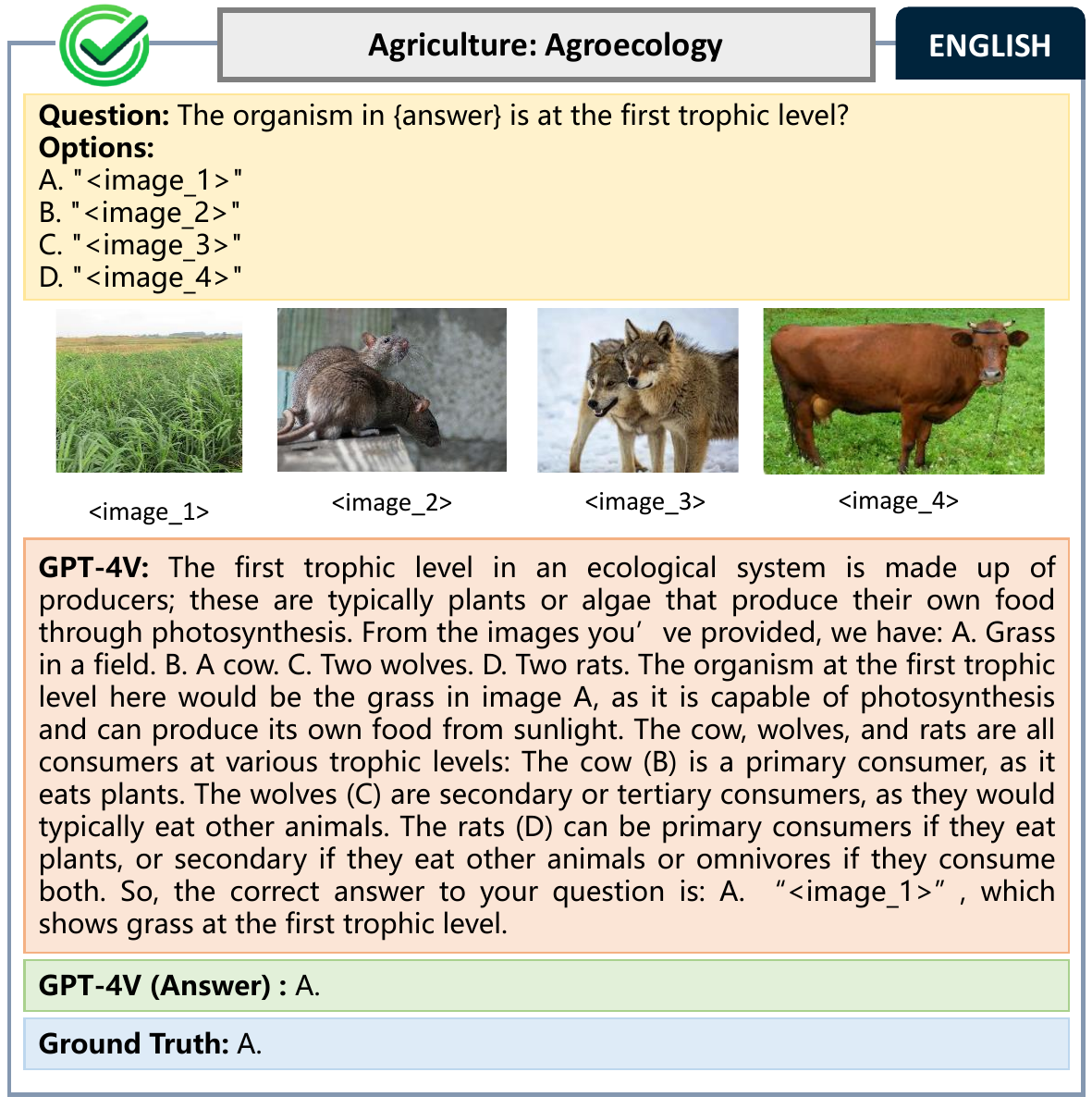}
    \caption{A sample correct case (subject: agriculture, language: English).}
    \label{fig:case_true_en}
\end{figure}
\begin{figure}[h]
    \centering
    \includegraphics[width=0.5\textwidth]{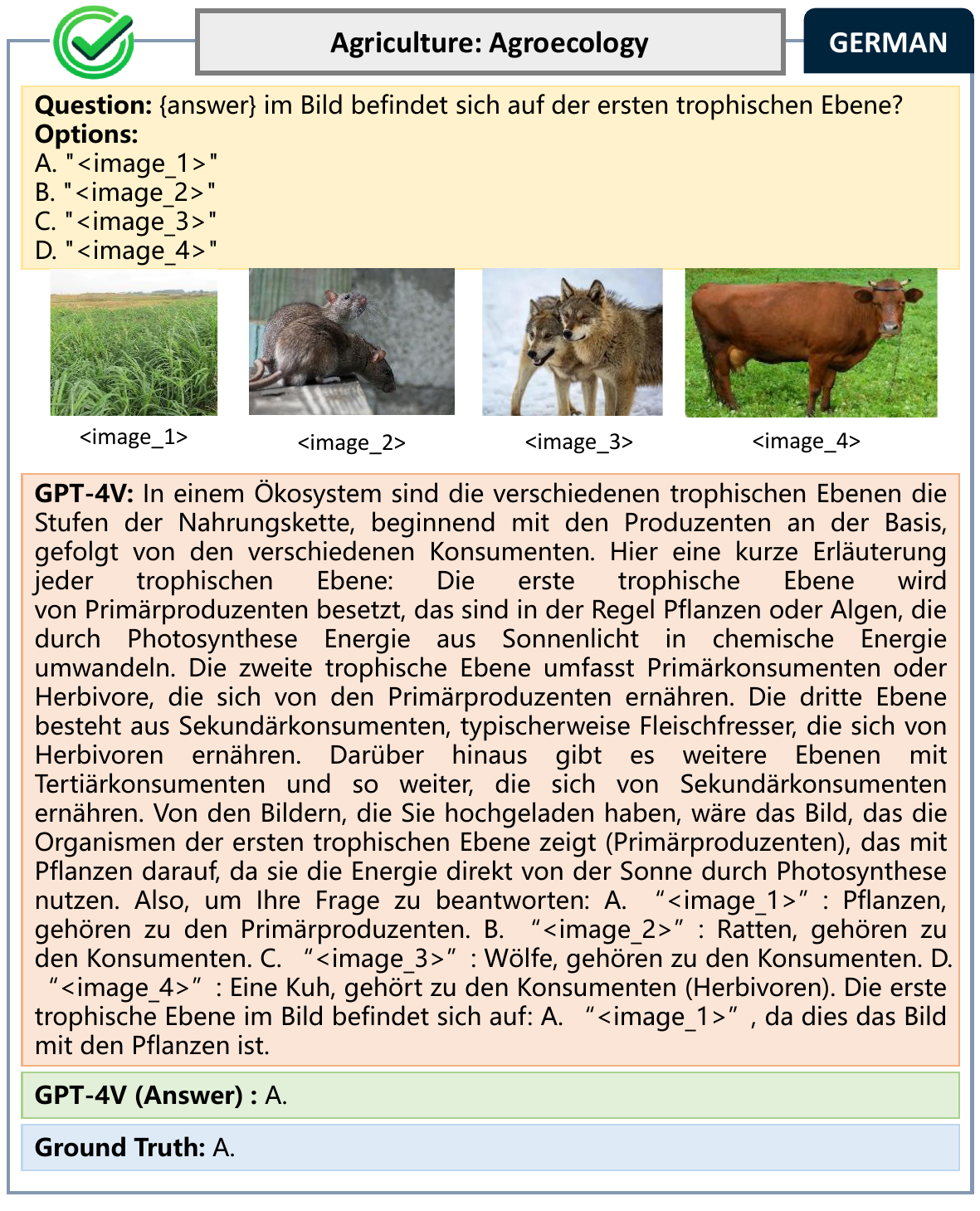}
    \caption{A sample correct case (subject: agriculture, language: German).}
    \label{fig:case_true_de}
\end{figure}
\begin{figure}[h]
    \centering
    \includegraphics[width=0.5\textwidth]{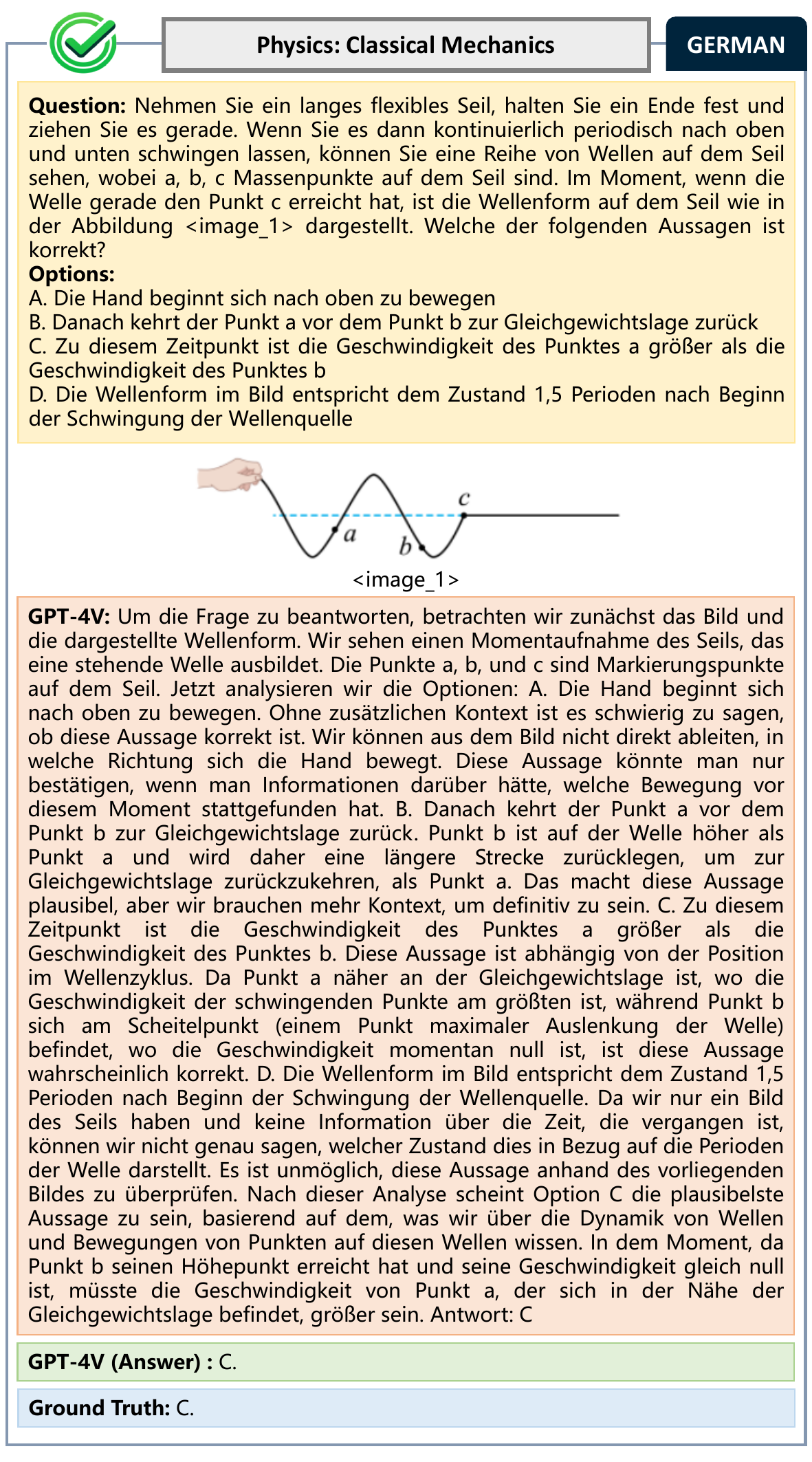}
    \caption{A sample correct case (subject: physics, language: German).}
    \label{fig:case2_de}
\end{figure}
\begin{figure}[h]
    \centering
    \includegraphics[width=0.5\textwidth]{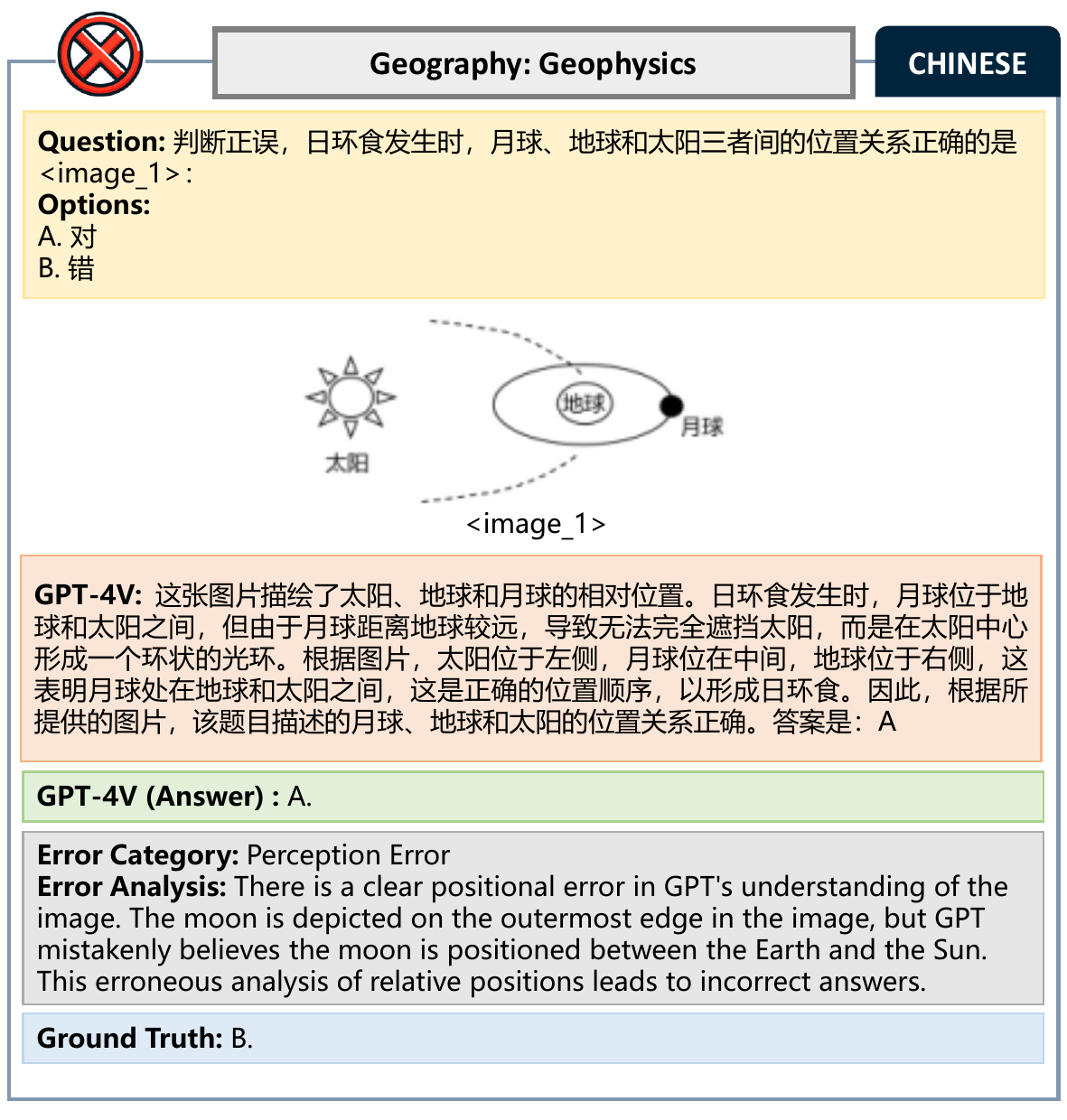}
    \caption{A sample perceptual error case (subject: geography, language: Chinese).}
    \label{fig:case8_cn}
\end{figure}
\begin{figure}[h]
    \centering
    \includegraphics[width=0.5\textwidth]{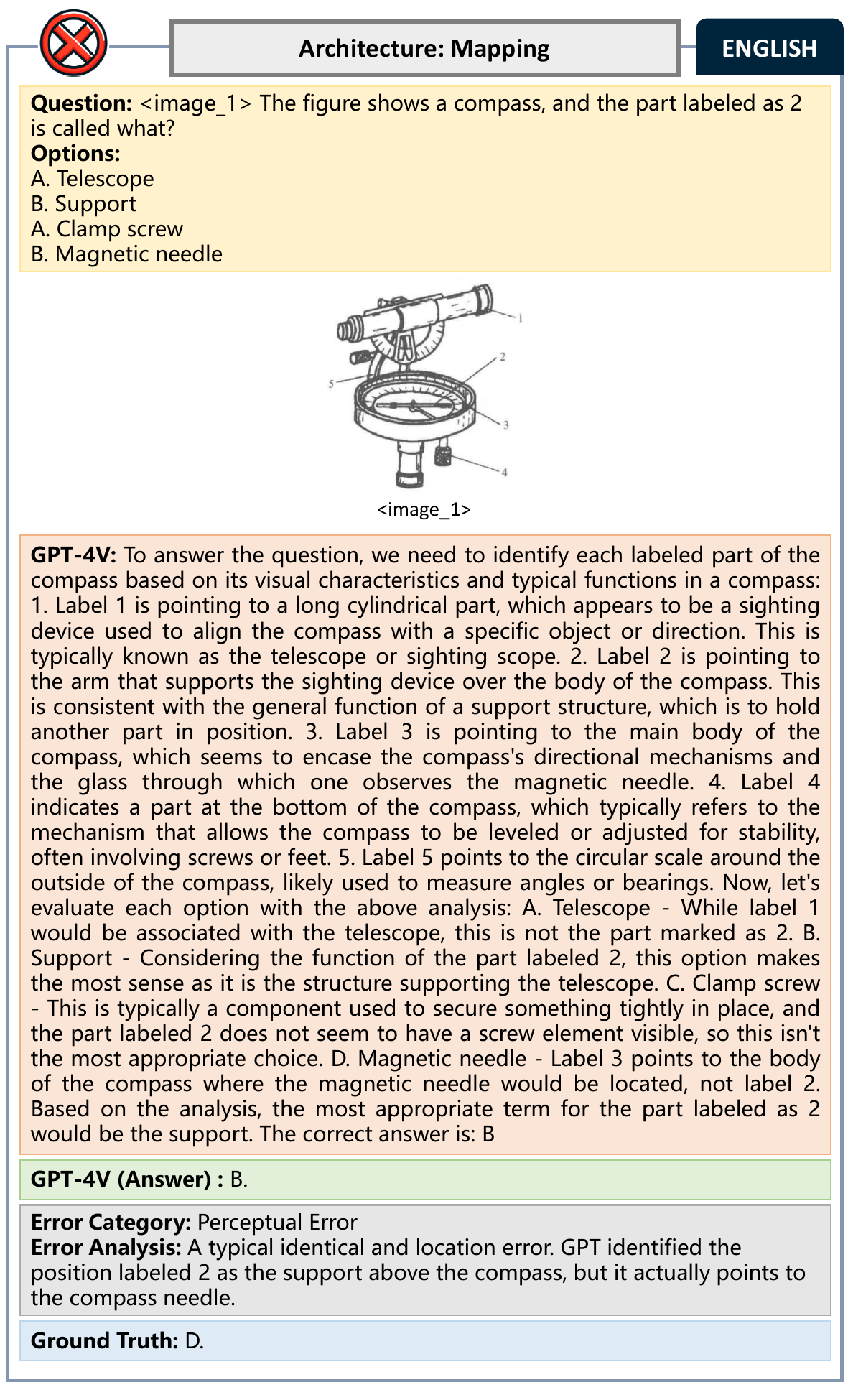}
    \caption{A sample perceptual error case (subject: architecture, language: English).}
    \label{fig:case9_en}
\end{figure}
\begin{figure}[h]
    \centering
    \includegraphics[width=0.5\textwidth]{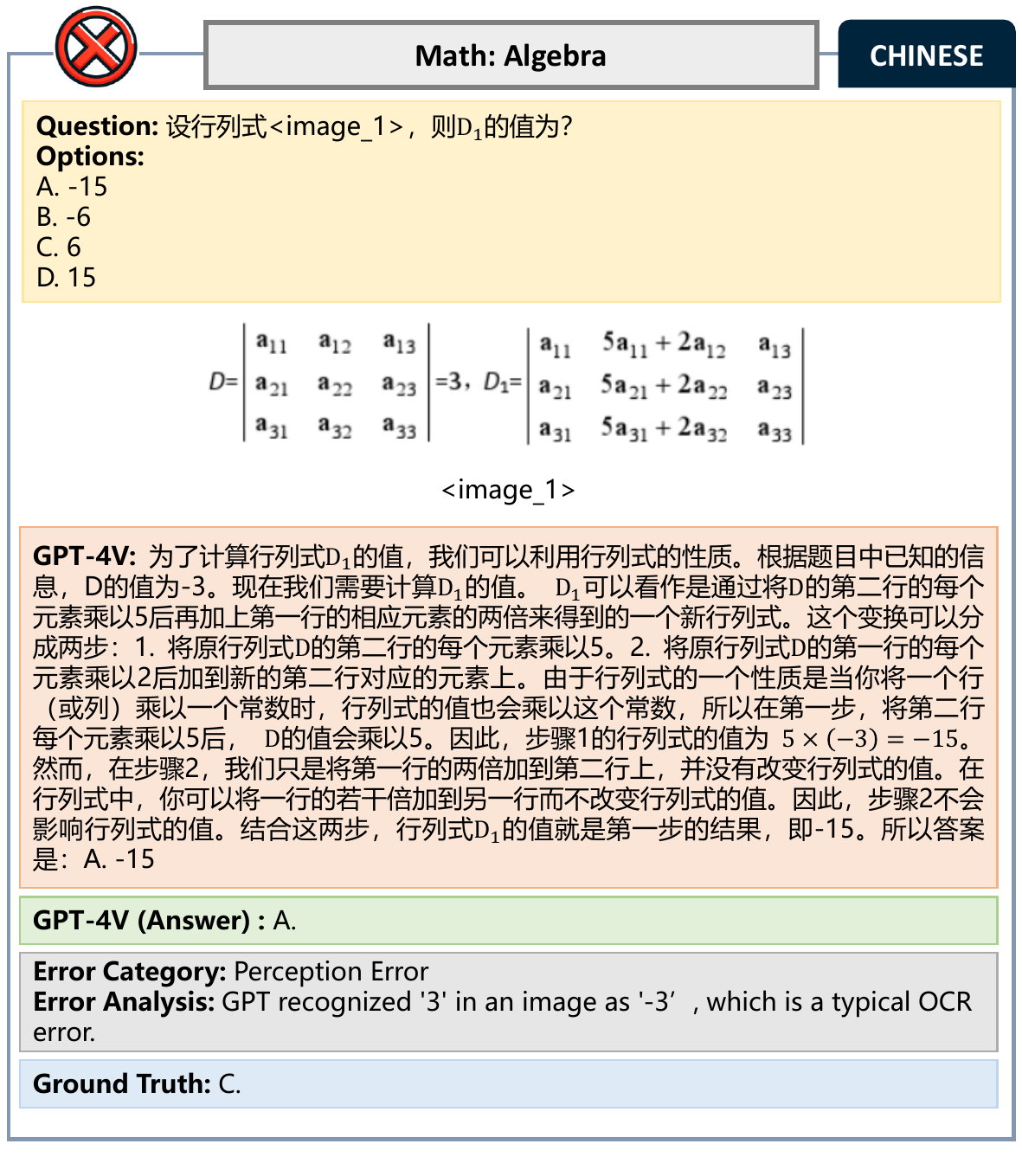}
    \caption{A sample perceptual error case (subject: math, language: Chinese).}
    \label{fig:case1_cn}
\end{figure}
\begin{figure}[h]
    \centering
    \includegraphics[width=0.5\textwidth]{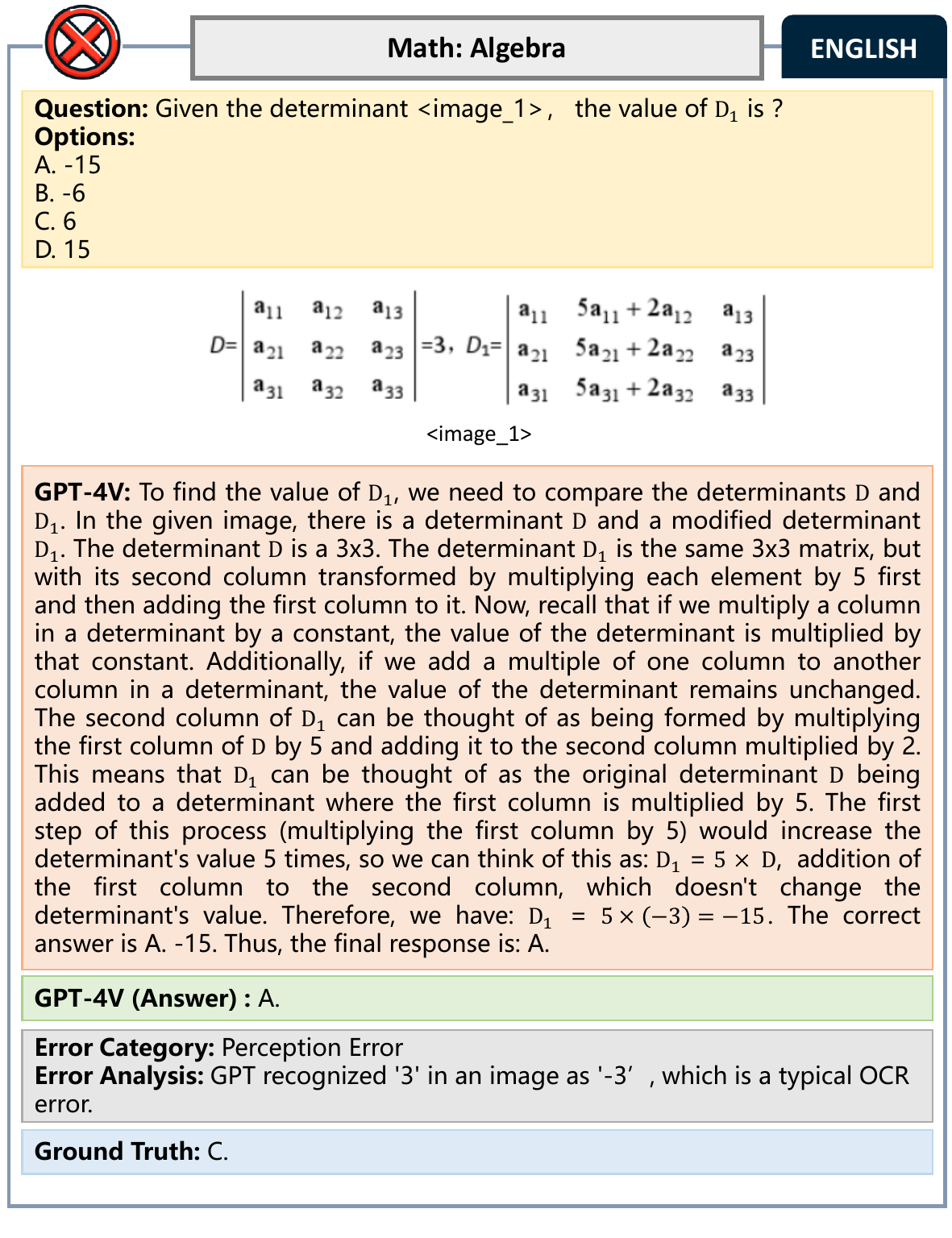}
    \caption{A sample perceptual error case (subject: math, language: English).}
    \label{fig:case1_en}
\end{figure}
\begin{figure}[h]
    \centering
    \includegraphics[width=0.5\textwidth]{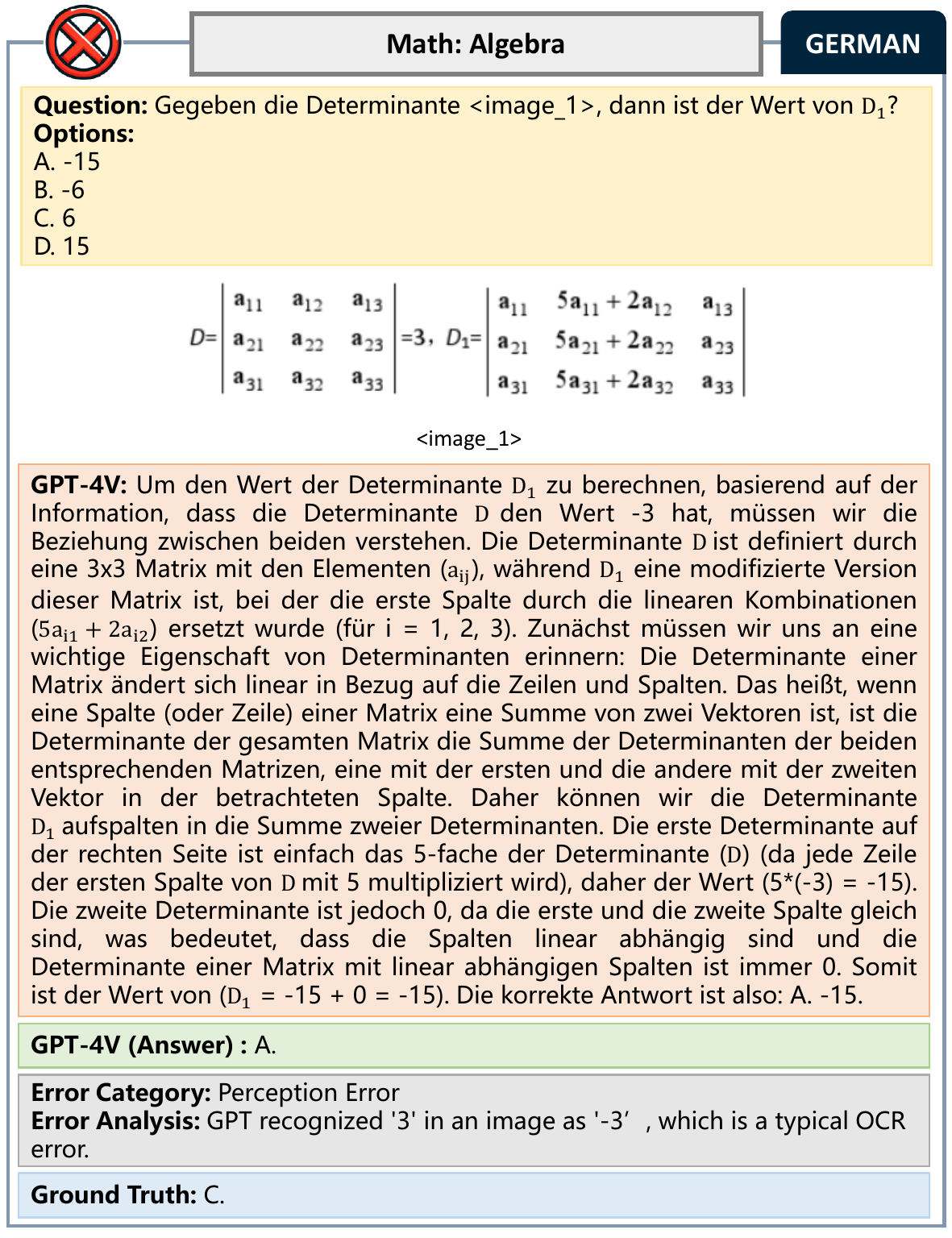}
    \caption{A sample perceptual error case (subject: math, language: German).}
    \label{fig:case1_de}
\end{figure}
\begin{figure}[h]
    \centering
    \includegraphics[width=0.5\textwidth]{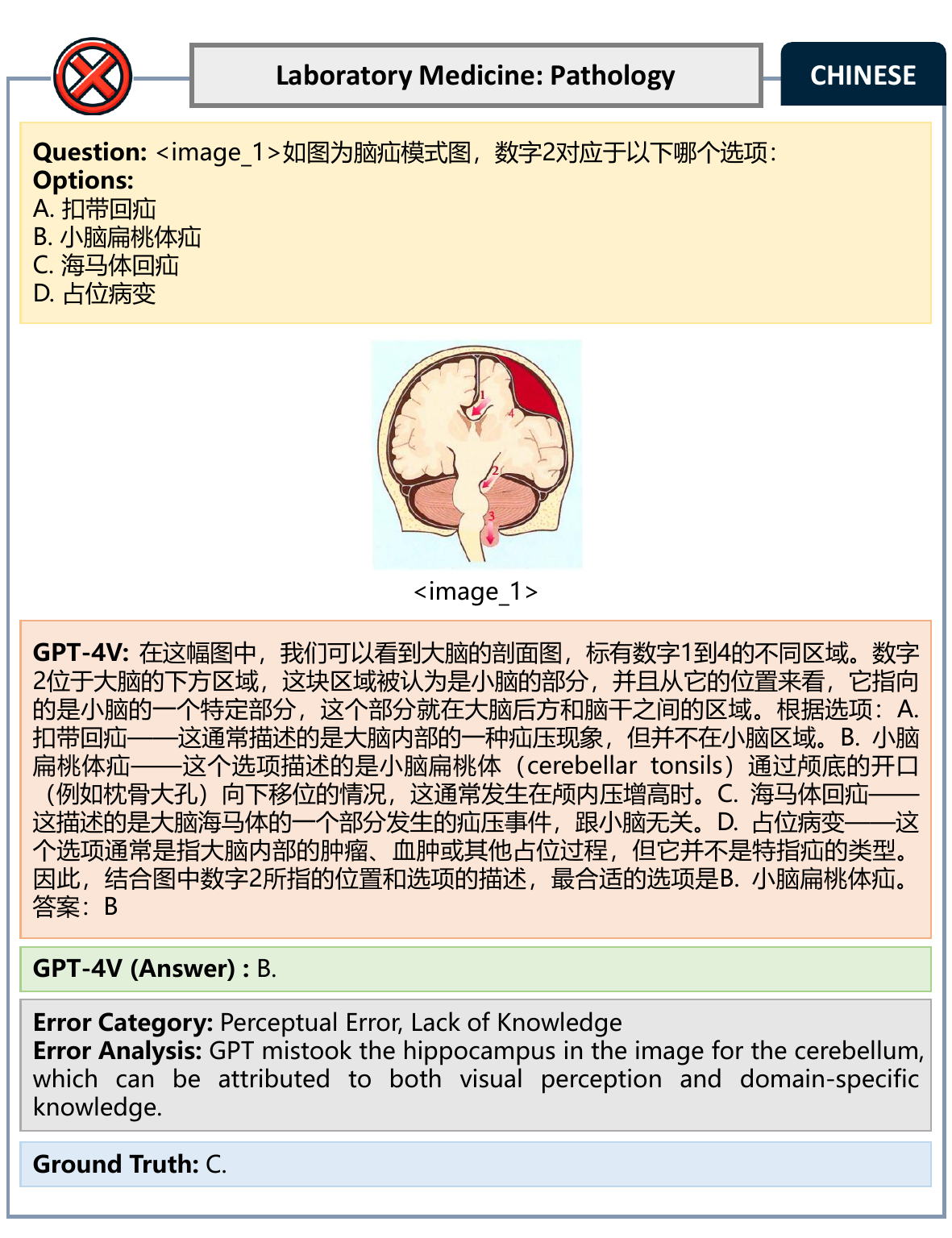}
    \caption{A sample perceptual error case (subject: laboratory medicine, language: Chinese).}
    \label{fig:case3_cn}
\end{figure}
\begin{figure}[h]
    \centering
    \includegraphics[width=0.5\textwidth]{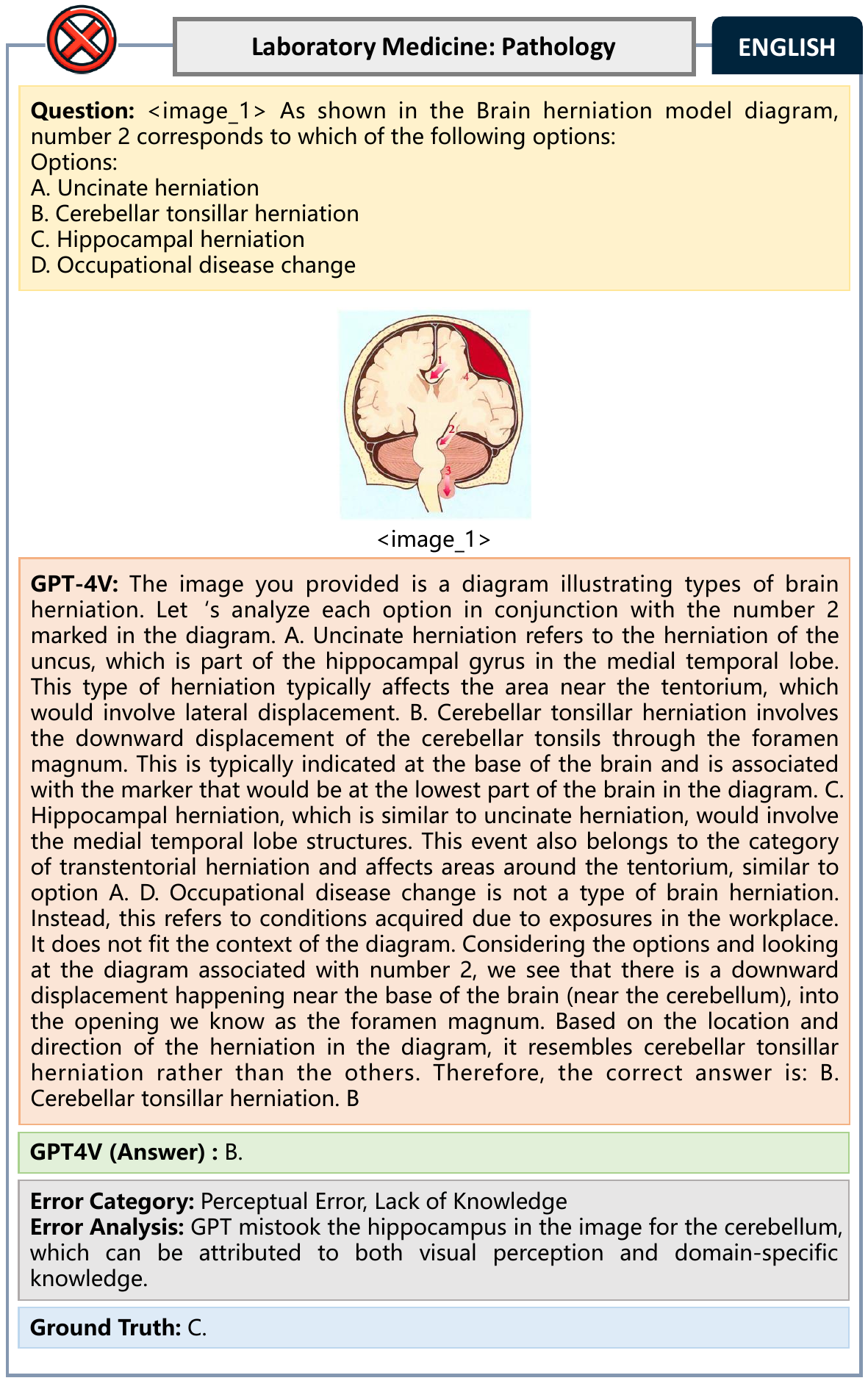}
    \caption{A sample perceptual error case (subject: laboratory medicine, language: English).}
    \label{fig:case3_en}
\end{figure}
\begin{figure}[h]
    \centering
    \includegraphics[width=0.5\textwidth]{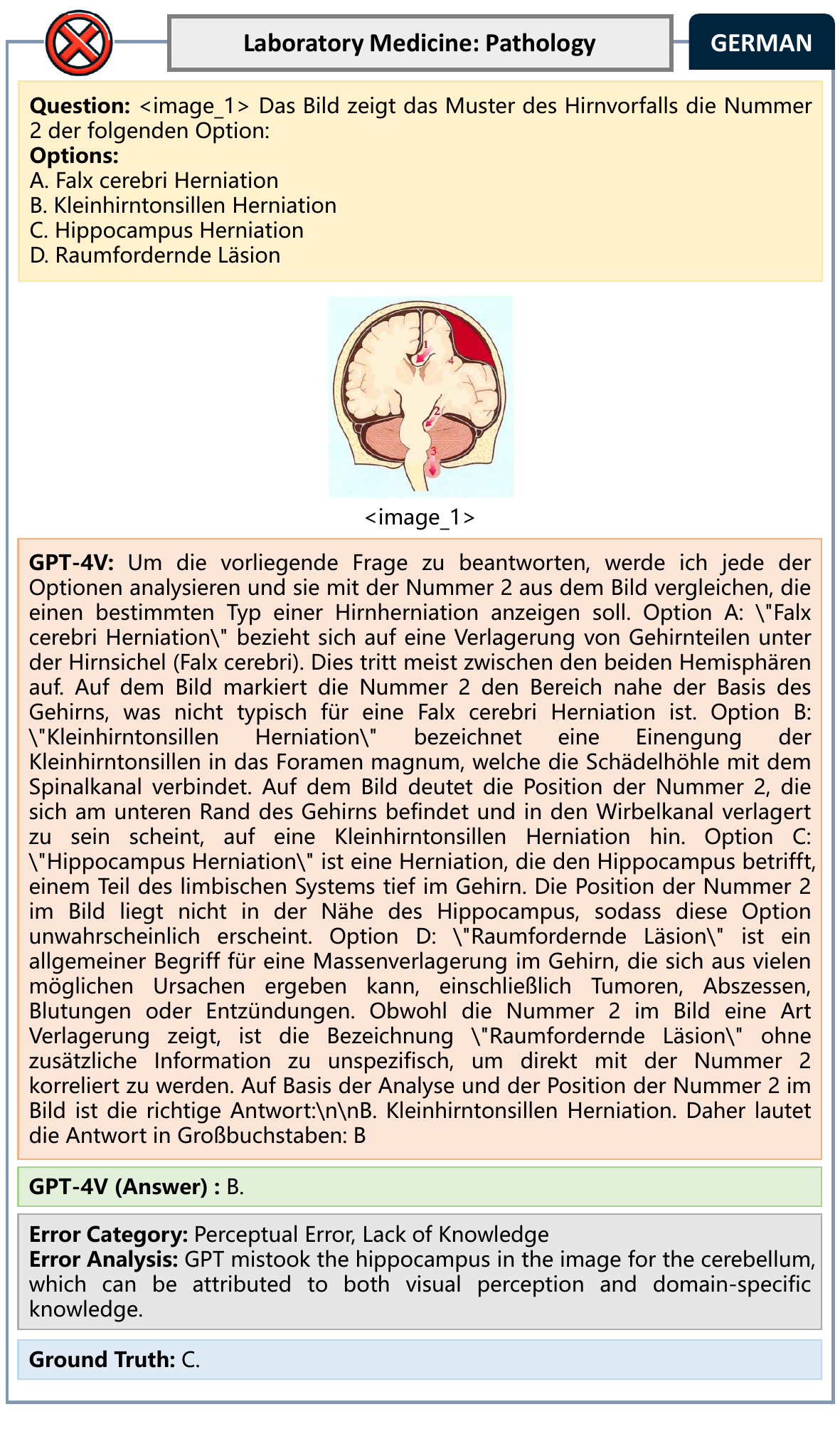}
    \caption{A sample perceptual error case (subject: laboratory medicine, language: German).}
    \label{fig:case3_de}
\end{figure}
\begin{figure}[h]
    \centering
    \includegraphics[width=0.5\textwidth]{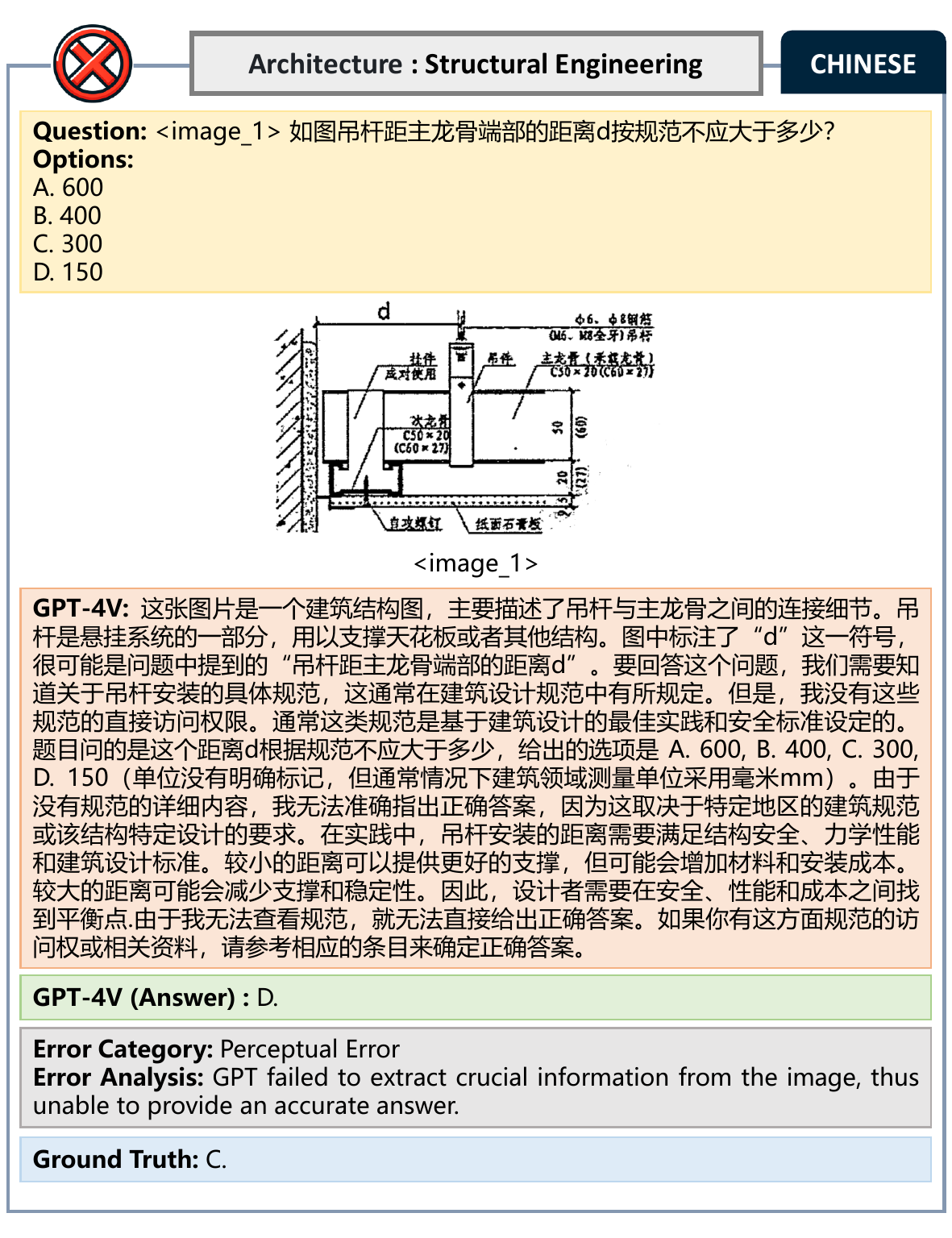}
    \caption{A sample perceptual error case (subject: architecture, language: Chinese).}
    \label{fig:case4_cn}
\end{figure}
\begin{figure}[h]
    \centering
    \includegraphics[width=0.5\textwidth]{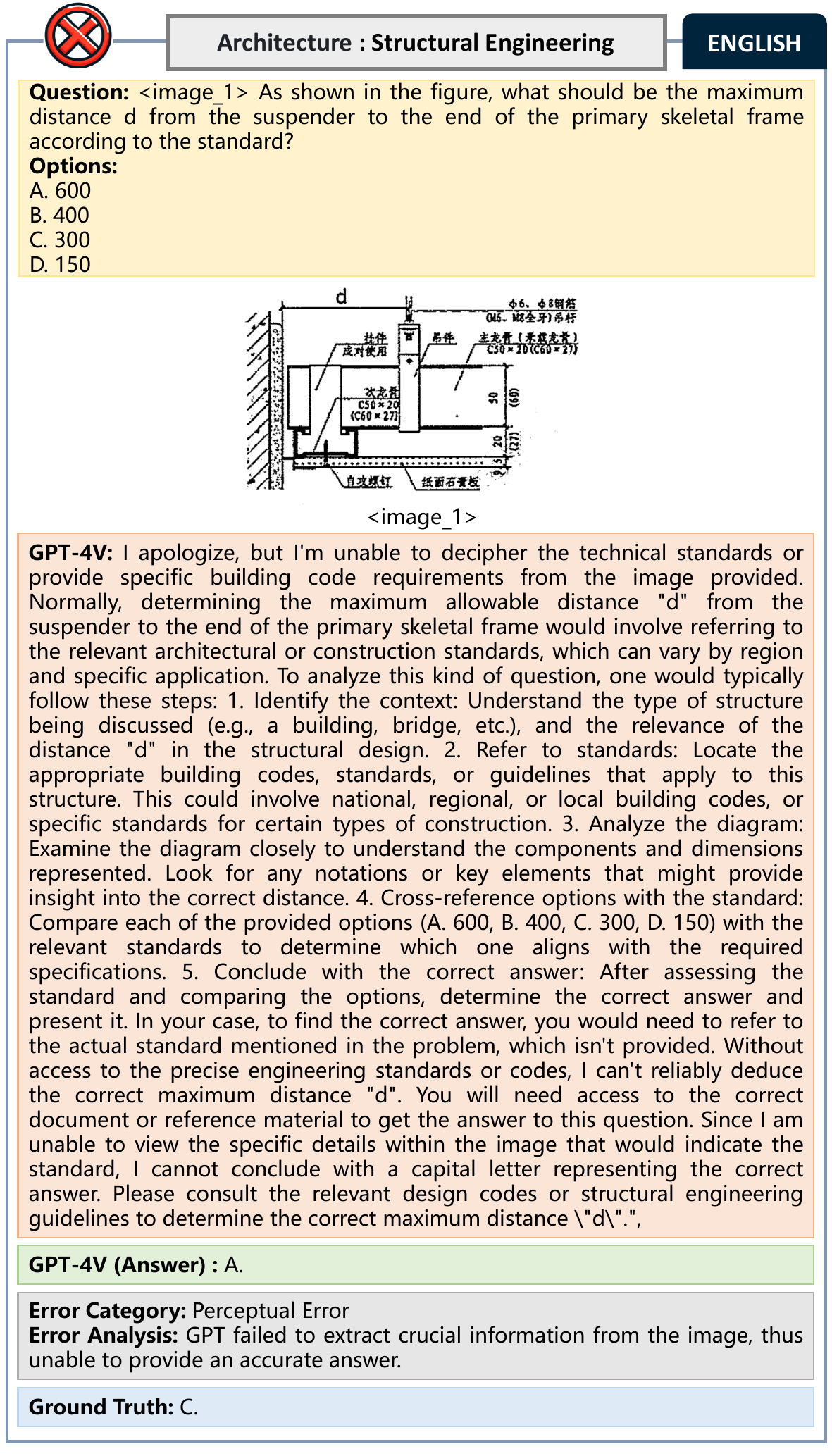}
    \caption{A sample perceptual error case (subject: architecture, language: English).}
    \label{fig:case4_en}
\end{figure}
\begin{figure}[h]
    \centering
    \includegraphics[width=0.5\textwidth]{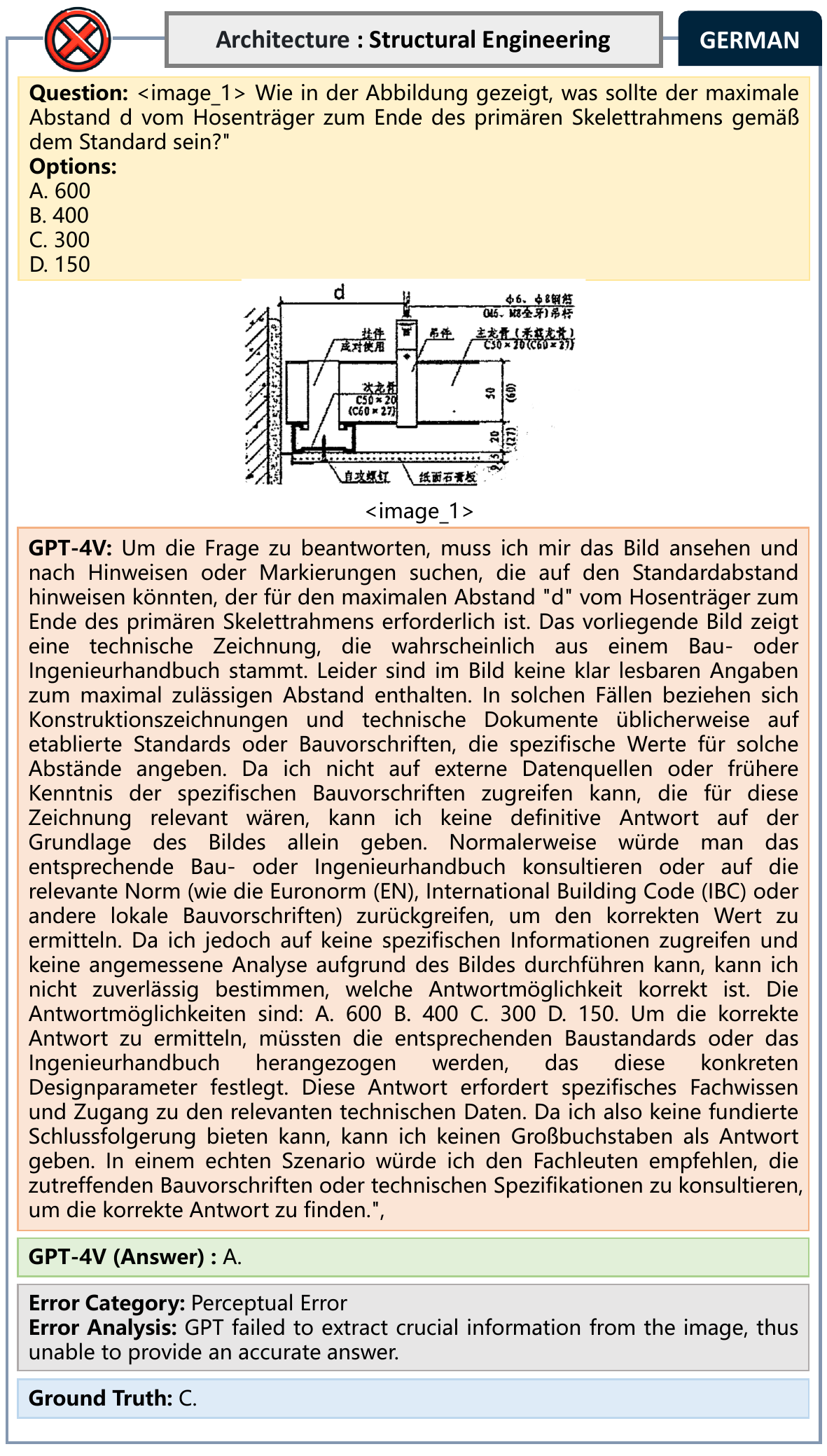}
    \caption{A sample perceptual error case (subject: architecture, language: German).}
    \label{fig:case4_de}
\end{figure}
\begin{figure}[h]
    \centering
    \includegraphics[width=0.5\textwidth]{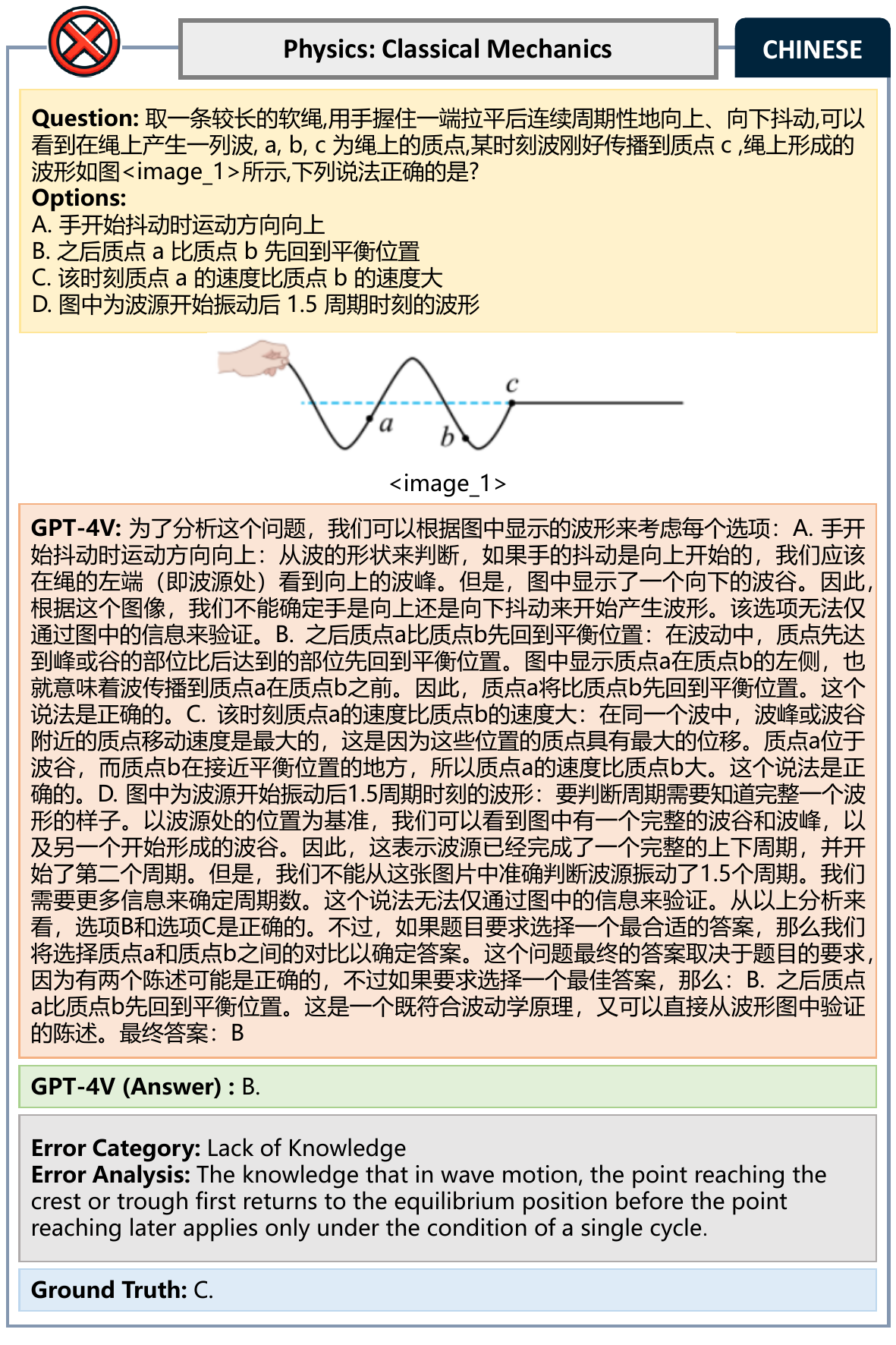}
    \caption{A sample lack of knowledge case (subject: physics, language: Chinese).}
    \label{fig:case2_cn}
\end{figure}
\begin{figure}[h]
    \centering
    \includegraphics[width=0.5\textwidth]{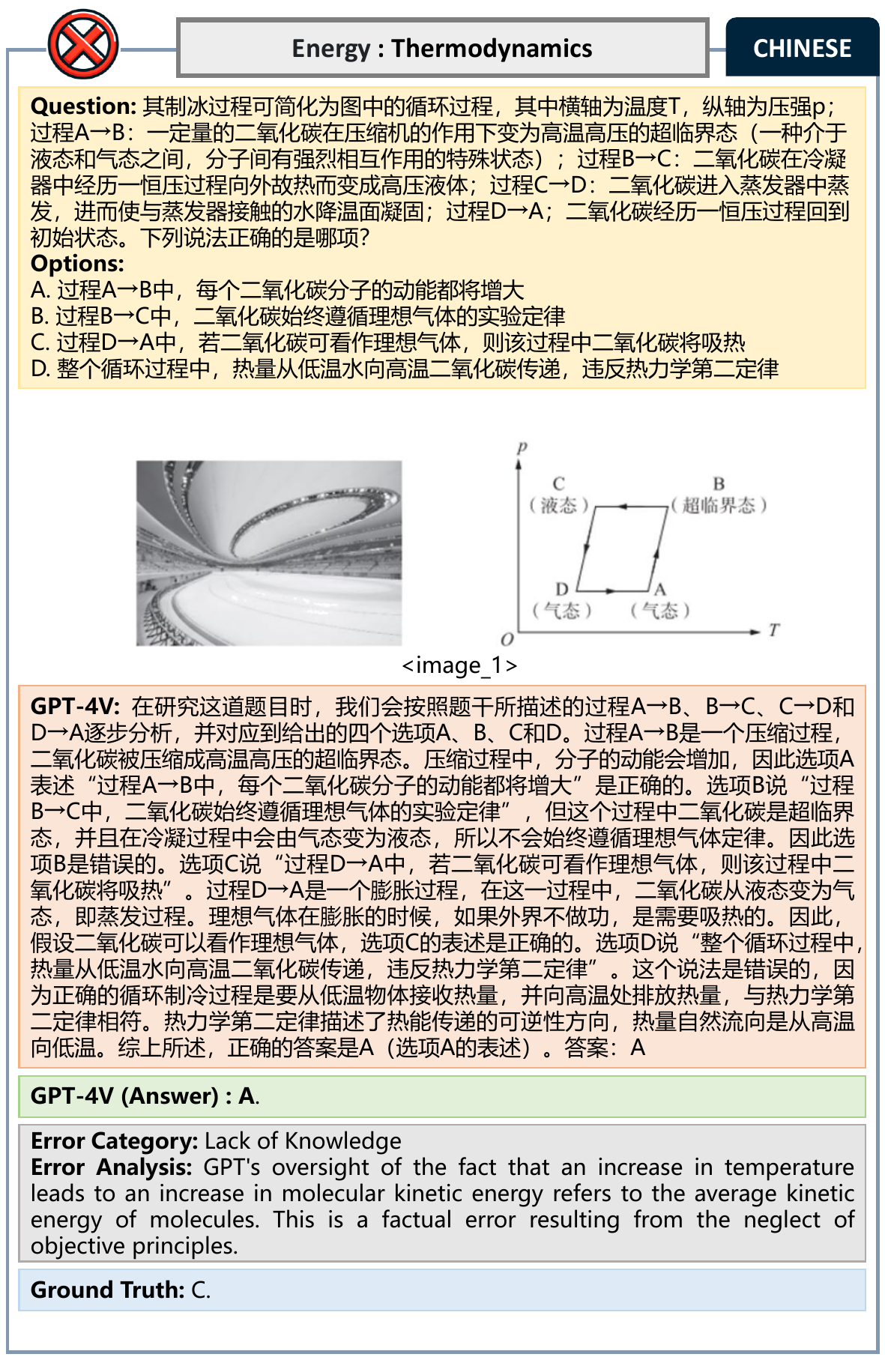}
    \caption{A sample lack of knowledge case (subject: energy, language: Chinese).}
    \label{fig:case5_cn}
\end{figure}
\begin{figure}[h]
    \centering
    \includegraphics[width=0.5\textwidth]{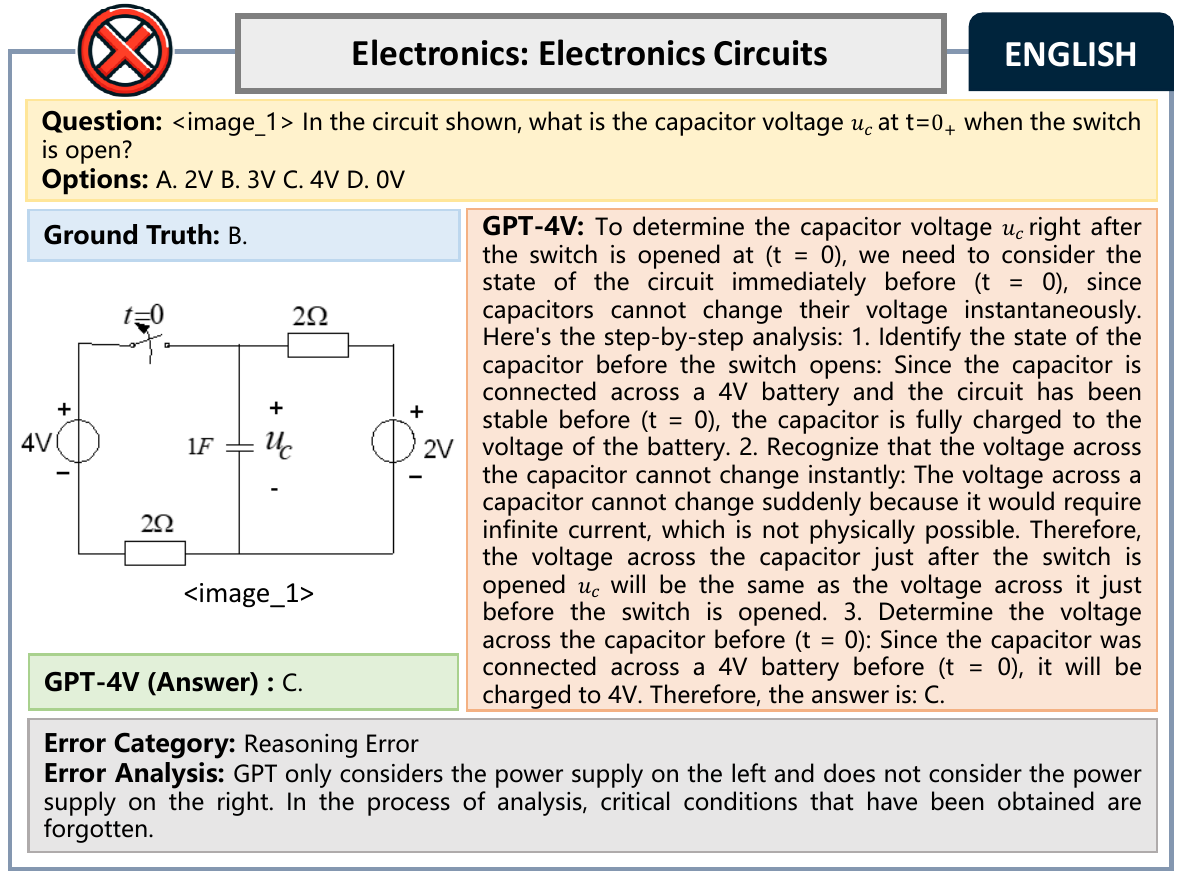}
    \caption{A typical case of reasoning error (subject: electronics, language: English).}
    \label{fig:case7_en}
\end{figure}
\begin{figure}[h]
    \centering
    \includegraphics[width=0.5\textwidth]{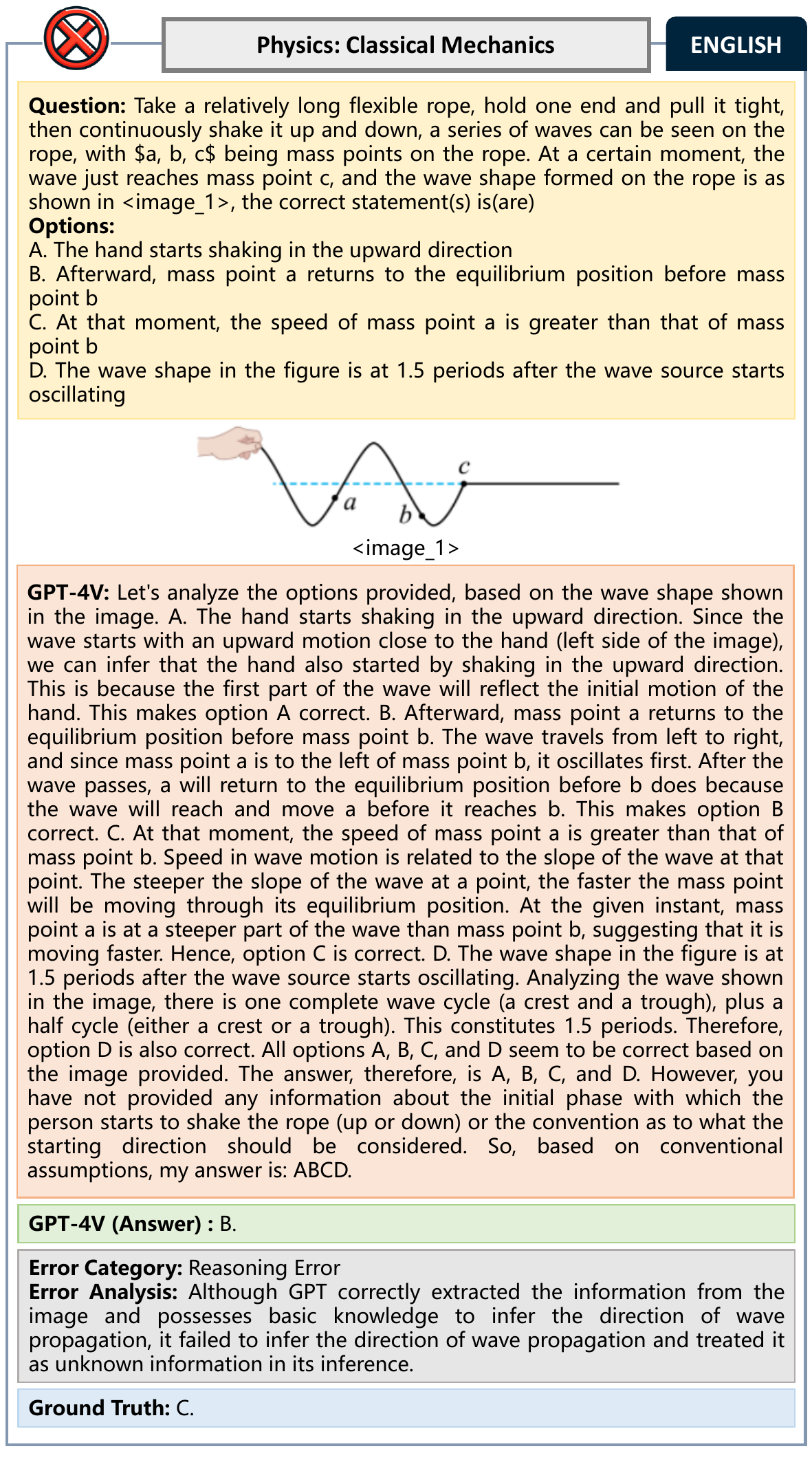}
    \caption{A sample reasoning error case (subject: physics, language: English).}
    \label{fig:case2_en}
\end{figure}
\begin{figure}[h]
    \centering
    \includegraphics[width=0.5\textwidth]{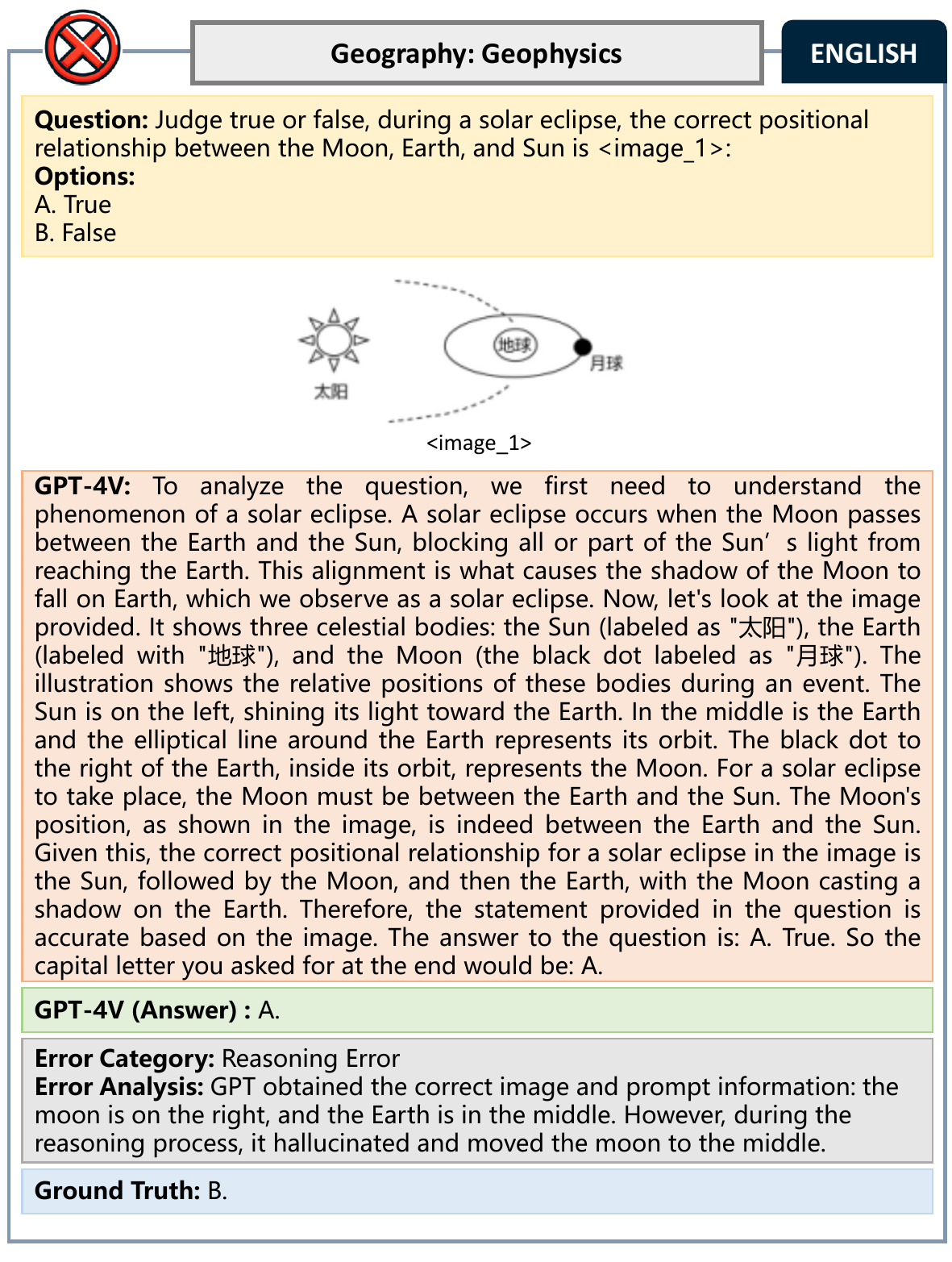}
    \caption{A sample reasoning error case (subject: geography, language: English).}
    \label{fig:case8_en}
\end{figure}
\begin{figure}[h]
    \centering
    \includegraphics[width=0.5\textwidth]{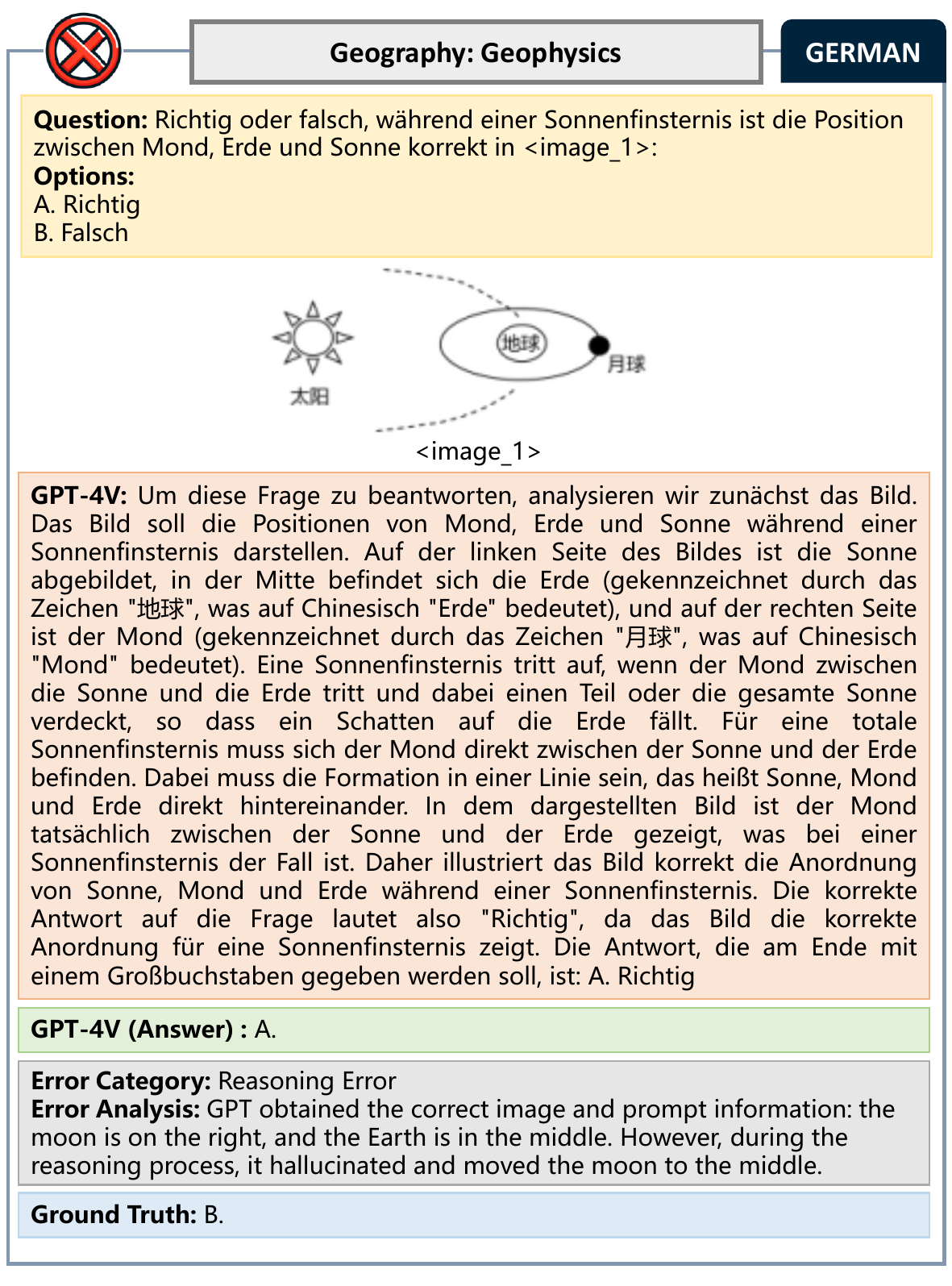}
    \caption{A sample reasoning error case (subject: geography, language: German).}
    \label{fig:case8_de}
\end{figure}

\vfill

\end{document}